\documentclass[10pt,twocolumn,letterpaper]{article}

\usepackage[pagenumbers]{cvpr} 

\usepackage[dvipsnames]{xcolor}

\definecolor{cvprblue}{rgb}{0.21,0.49,0.74}
\usepackage[pagebackref,breaklinks,colorlinks,citecolor=cvprblue]{hyperref}

\usepackage[utf8]{inputenc} 
\usepackage[T1]{fontenc}    
\usepackage{hyperref}       
\usepackage{url}            
\usepackage{booktabs}       
\usepackage{amsfonts}       
\usepackage{nicefrac}       
\usepackage{microtype}      
\usepackage{xcolor}         
\usepackage{comment}
\usepackage{amsmath}
\usepackage{multirow}
\usepackage{adjustbox}
\usepackage{color}
\usepackage{caption}
\usepackage{bm}
\usepackage{enumitem}
\usepackage{float}
\usepackage{xcolor,colortbl}
\usepackage{wrapfig}
\usepackage{subcaption}

\newcommand\sotaa{\textcolor{red}}
\newcommand\sotab{\textcolor{blue}}
\definecolor{Gray}{rgb}{0.95, 0.95, 0.95}
\newcolumntype{a}{>{\columncolor{Gray}}c}

\renewcommand{\thefootnote}{\fnsymbol{footnote}}

\begin{document}

\title{Transcending the Limit of Local Window: \\
Advanced Super-Resolution Transformer with Adaptive Token Dictionary}

\author{Leheng Zhang$^{1}$ \quad Yawei Li$^{2,3}$ \quad Xingyu Zhou$^1$ \quad Xiaorui Zhao$^1$ \quad Shuhang Gu$^{1}$\footnotemark[1]\\
$^1$University of Electronic Science and Technology of China \quad $^2$Computer Vision Lab, ETH Zürich\\
$^3$Integrated Systems Laboratory, ETH Zürich\\
{\tt \small \{lehengzhang12, shuhanggu\}@gmail.com} \\
{\small \url{https://github.com/LabShuHangGU/Adaptive-Token-Dictionary}}
}

\twocolumn[{%
\renewcommand\twocolumn[1][]{#1}%
\maketitle
\begin{center}

    \vspace{-4mm}

    \centering
    \captionsetup{type=figure}
    \includegraphics[width=\textwidth]{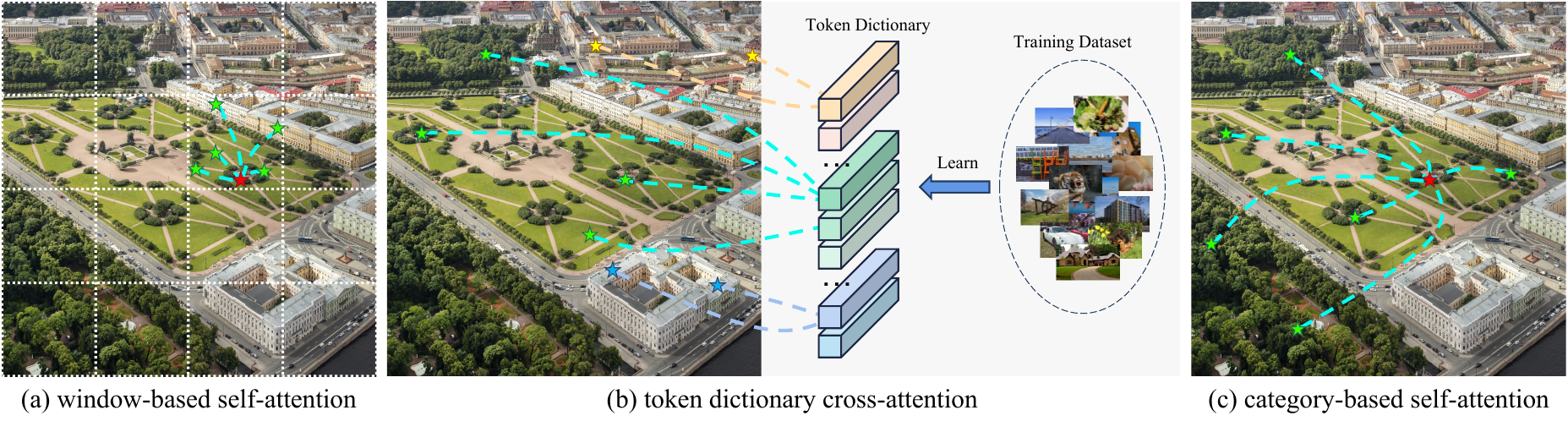}
    
    \vspace{-4mm}
    
    \captionof{figure}{
    Three different kinds of attention mechanism: (a) window-based self-attention exploits tokens in the same local window to enhance image tokens; (b) our proposed token dictionary cross-attention leverages the auxiliary dictionary to summarize and incorporate global information to the image tokens; (c) our proposed category-based self-attention adopts category labels to divide image tokens.
    }
    
\end{center}%
}
]

\renewcommand{\thefootnote}{\fnsymbol{footnote}}
\footnotetext[1]{corresponding author}

\vspace{-2mm}

\begin{abstract}

\vspace{-1mm}

Single Image Super-Resolution is a classic computer vision problem that involves estimating high-resolution (HR) images from low-resolution (LR) ones.
Although deep neural networks (DNNs), especially Transformers for super-resolution, have seen significant advancements in recent years, challenges still remain, particularly in limited receptive field caused by window-based self-attention.
To address these issues, we introduce a group of auxiliary \textbf{A}daptive \textbf{T}oken \textbf{D}ictionary to SR Transformer and establish an \textbf{ATD}-SR method.
The introduced token dictionary could learn prior information from training data and adapt the learned prior to specific testing image through an adaptive refinement step.
The refinement strategy could not only provide global information to all input tokens but also group image tokens into categories.
Based on category partitions, we further propose a category-based self-attention mechanism designed to leverage distant but similar tokens for enhancing input features.
The experimental results show that our method achieves the best performance on various single image super-resolution benchmarks.
\end{abstract}

\vspace{-2mm}

\section{Introduction}

\vspace{1mm}

The task of single image super-resolution (SR) aims to recover clean high-quality (HR) images from a solitary degraded low-quality (LR) image.
Since each LR image may correspond to a mass of possible HR counterparts, image SR is a classical ill-posed and challenging problem in the fields of computer vision and image processing.
This practice is significant as it transcends the resolution and accuracy limitations of cost-effective sensors and improves images produced by outdated equipment. 

The evolution of image super-resolution techniques has shifted from earlier methods like Markov random fields~\cite{he2011single} and Dictionary Learning~\cite{yang2010image} to advanced deep learning approaches. The rise of deep neural networks, particularly convolutional neural networks (CNNs), marked a significant improvement in this field, with models effectively learning mapping functions from LR to HR images~~\cite{Dong_2015_srcnn, Kim_2016_vdsr, lim2017edsr, zhang2018rcan, Dai_2020_san}. More recently, Transformer-based neural networks have outperformed CNNs in image super-resolution by employing self-attention mechanisms to better model long-range image structures~\cite{liang2021swinir, chen2023activating, li2023grl}.

Despite recent advances in image SR, several challenges remain unresolved.
One major issue faced by SR transformers is the balancing act between achieving satisfactory SR accuracy and managing increased computational complexity.
Due to the quadratic computational complexity of the self-attention mechanism, previous methods~\cite{liu2021swin, liang2021swinir} have been forced to confine attention computation to local windows to manage computational load.
However, this window-based method imposes a constraint on the receptive field, affecting the performance.
Although recent studies~\cite{chen2023activating, li2023grl} indicate that expanding window size improves the receptive field and enhances SR performance, it exacerbates the curse of dimensionality.
This issue underscores the need for an efficient method to effectively model long-range dependencies, without being constrained within local windows.
Furthermore, conventional image SR often employs general-purpose computations that do not take the content of the image into account.
Rather than employing a partitioning strategy based on rectangular local windows, opting for division according to the specific content categories of an image could be more beneficial to the SR process.

In our paper, we draw inspiration from classical dictionary learning in super-resolution to introduce a token dictionary, enhancing both cross- and self-attention calculations in image processing. This token dictionary offers three distinct benefits.
\textbf{Firstly}, it enables the use of cross-attention to integrate external priors into image analysis. 
This is achieved by learning auxiliary tokens that encapsulate common image structures, facilitating efficient processing with linear complexity in proportion to the image size.
\textbf{Secondly}, it enables the use of global information to establish long-range connection. 
This is achieved by refining the dictionary with activated tokens to summarize image-specific information globally through a reversed form of attention.
\textbf{Lastly}, it enables the use of all similar parts of the image to enhance image tokens without being limited by local window partitions.
This is achieved by content-dependent structural partitioning according to the similarities between image and dictionary tokens for category-based self-attention.
These innovations enable our method to significantly outperform existing state-of-the-art techniques without substantially increasing the model complexity.

 Our contributions can be summarized as follows:
 \begin{itemize}

     \item We introduce the idea of token dictionary, which utilizes a group of auxiliary tokens to provide prior information to each image token and summarize prior information from the whole image, effectively and efficiently in a cross-attention manner.

     \item We exploit our token dictionary to group image tokens into categories and break through boundaries of local windows to exploit long-range prior in a category-based self-attention manner.
     
     \item By combining the proposed token dictionary cross-attention and category-based self-attention, our model could leverage long-range dependencies effectively and achieve superior super-resolution results over the existing state-of-the-art methods.
    
 \end{itemize}

\section{Related Works}
\label{related_works}

The past decade has witnessed numerous endeavors aimed at improving the performance of deep learning methods across diverse fields, including the single image super-resolution.
Pioneered by SRCNN~\cite{Dong_2015_srcnn}, which introduces deep learning to super-resolution with a straightforward 3-layer convolutional neural network (CNN), 
numerous studies have since explored various architectural enhancements to boost performance~\cite{Kim_2016_vdsr, lim2017edsr, Zhang_2018_rdn, zhang2018rcan, Dai_2020_san, Niu_2020_han, Mei_2021_nlsa, Mei2020image, kim2016deeply}.
VDSR~\cite{Kim_2016_vdsr} implements a deeper network, and DRCN~\cite{kim2016deeply} proposes a recursive structure.
EDSR~\cite{lim2017edsr} and RDN~\cite{Zhang_2018_rdn} develop new residual blocks, further improving CNN capability in SR.
Drawing inspiration from Transformer~\cite{vaswani2017attention}, \citet{wang2018non} first integrates non-local attention block into CNN, validating the effects of attention mechanism in vision tasks.
Following that, numerous advances in attention have emerged.
CSNLN~\cite{Mei2020image} makes use of non-local cross-scale attention to explore cross-scale feature correlations and mine self-exemplars in natural images.
RCAN~\cite{zhang2018rcan} and SAN~\cite{Dai_2020_san} respectively incorporate channel attention to capture interdependencies between different channels.
NLSA~\cite{Mei_2021_nlsa} further improves efficiency through sparse attention, which reduces the calculation between unrelated or noisy contents.

Recently, with the introduction of ViT~\cite{Dosovitskiy_2020_vit} and its variants~\cite{liu2021swin, Chu_2021_twin, Wang_2022_pvt}, the efficacy of pure Transformer-based models in image classification has been established. 
Based on this, IPT~\cite{Chen_2020_ipt} makes a successful attempt to exploit the Transformer-based network for various image restoration tasks. 
Since then, a variety of techniques have been developed to enhance the performance of super-resolution transformers. 
This includes the implementation of shifted window self-attention by SwinIR~\cite{liang2021swinir} and CAT~\cite{chen2022cross}, group-wise multi-scale self-attention by ELAN~\cite{zhang2022efficient}, sparse attention by ART~\cite{zhang2023accurate} and OmniSR~\cite{omni_sr}, anchored self-attention by GRL~\cite{li2023grl}, and more, all aimed at expanding the scope of receptive field to achieve better results.
Furthermore, strategies such as pretraining on extensive datasets~\cite{li2021efficient}, employing ConvFFN~\cite{omni_sr}, and utilizing large window sizes~\cite{chen2023activating} have been employed to boost performance, indicating the growing adaptability and impact of Transformer-based approaches in the field of image SR.

In this paper, building upon the effectiveness of the attention mechanism in image SR, we propose two types of attention: token dictionary cross-attention (TDCA) to leverage external prior and adaptive category-based multi-head self-attention (AC-MSA) to model long-range dependencies.
When synergized with window-based attention, our approach seamlessly integrates local, global, and external information, yielding promising outcomes in image super-resolution tasks.

\section{Methodology}
\label{headings}

\subsection{Motivation}
\label{pre}
In this subsection, we introduce the motivation of our approach.
We first discuss how dictionary-learning-based SR methods utilize learned dictionary to provide supplementary information for image SR. 
Then, we analyze the attention operation and discuss its similarity to the coefficient calculation and signal reconstruction processes in dictionary-learning-based methods.
Lastly, we discuss how these two methods motivate us to introduce an auxiliary token dictionary for enhancing both cross- and self-attention calculations in image processing.

\noindent\textbf{Dictionary Learning for Image Super-Resolution. } 
Before the era of deep learning, dictionary learning plays an important role in providing prior information for image SR.
Due to the limited computational resources, conventional dictionary-learning-based methods divide image into patches for modeling image local prior.
Denote $\bm{x}\in \mathbb{R}^{d}$ as a vectorized image patch in the low-resolution (LR) image.
To estimate the corresponding high-resolution (HR) patch $\bm{y}\in \mathbb{R}^{d}$,
\citet{yang2010image} decompose the signal by solving the sparse representation problem:
\begin{equation}
    \label{eq:sr}
    \bm{\alpha^*} = argmin_{\bm{\alpha}}\|\bm{x} - \bm{D}_L \bm{\alpha}\|_2^2 + \lambda \|\bm{\alpha}\|_1
\end{equation}
and reconstruct the HR patch with $\bm{D}_{H}\bm{\alpha^*}$; where $\bm{D}_{L}\in \mathbb{R}^{d\times M}$ and $\bm{D}_H\in \mathbb{R}^{d\times M}$ are the learned LR and HR dictionaries, and $M$ is the number of atoms in the dictionary.
Most of dictionary-learning-based SR methods~\cite{yang2010image,Zeyde_2012_set14} learn coupled dictionaries $\bm{D}_{L}$ and $\bm{D}_{H}$ to summarize the prior information from the external training dataset;
several attempts~\cite{mairal2010online, mairal2008supervised} have also been made to refine dictionary according to the testing image for better SR results.

\noindent\textbf{Vision Transformer for Image Super-Resolution.} 
Recently, Transformer-based approaches have pushed the state-of-the-art of many vision tasks to a new level.
At the core of Transformer is the self-attention operation, which exploits similarity between tokens as weight to mutually enhance image features:
%
\begin{equation}
    \label{eq:sa}
    \begin{array}{c}
        \operatorname{Attention}(\bm{Q}, \bm{K}, \bm{V}) = \operatorname{SoftMax}\left(\bm{Q}\bm{K}^T/\sqrt{d}\right)\bm{V};
    \end{array}
\end{equation}
$\bm{Q}\in \mathbb{R}^{N\times d}$, $\bm{K}\in \mathbb{R}^{N\times d}$ and $\bm{V}\in \mathbb{R}^{N\times d}$ are linearly transformed from the input feature $\bm{X}\in \mathbb{R}^{N\times d}$ itself, $N$ is the token number and $d$ is the feature dimension.
Due to the self-attentive processing philosophy, the large window size plays a critical role in modeling the internal prior of more patches.
However, the complexity of self-attention computation increases quadratically with the number of input tokens, and different strategies including shift-window~\cite{liu2021swin,liu2022swin,liang2021swinir,conde2023swin2sr}, anchor attention~\cite{li2023grl}, and shifted crossed attention~\cite{li2021efficient} have been proposed to alleviate the limited window size issue of the Vision Transformer.

\noindent\textbf{Advanced Cross\&Self-Attention with Token Dictionary.}
After reviewing the above content, we found that the decomposition and reconstruction idea of dictionary learning-based image SR is similar to the process of self-attention computation.
Specifically, the above method in \cref{eq:sr} solves the sparse representation model to find similar LR dictionary atoms and reconstruct HR signal with the corresponding HR dictionary atoms; while attention-based methods use normalized point product operation to determine attention weights to combine value tokens.

The above observation implies that the idea of dictionary learning can be easily incorporated into the Transformer framework for breaking the limit of local window.
Specifically, a similar idea of coupled dictionary learning can be adopted in a token dictionary learning manner.
In the following subsection~\cref{sec:TDCA}, we introduce how we establish a token dictionary to learn typical structures from the training dataset and utilize cross attention operation to provide the learned supplementary information to all the image tokens.
Moreover, inspired by the image-specific online dictionary learning approach~\cite{mairal2010online, mairal2008supervised}, we further propose an adaptive dictionary refinement strategy in subsection~\cref{sec:ADR}.
By refining the dictionary with activated tokens, we could adapt the learned external dictionary to image-specific dictionary to better fit the image content and propagate the summarized global information to all the image tokens.
Another advantage of the introduced token dictionary lies in its similarity matrix with the image tokens.
According to the indexes of the closest dictionary items, we are able to group image tokens into categories.
Instead of leveraging image tokens in the same local window to enhance image feature, the proposed category-based self-attention module (subsection~\cref{sec:AC-MSA}) allows us to take benefit from similar tokens from the whole image.

\begin{figure*}
  \centering
  \begin{subfigure}{.48\linewidth}
    \centering
    \includegraphics[width=\linewidth]{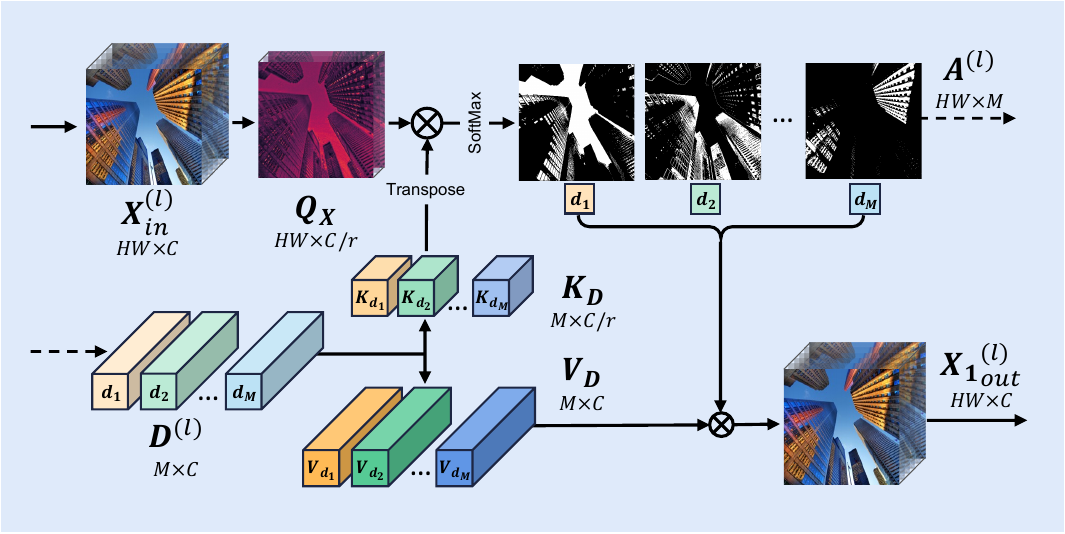}
    \caption{Token Dictionary Cross-Attention}
    \label{fig:tdca}
  \end{subfigure}%
  \hfill
  \begin{subfigure}{.48\linewidth}
    \centering
    \includegraphics[width=\linewidth]{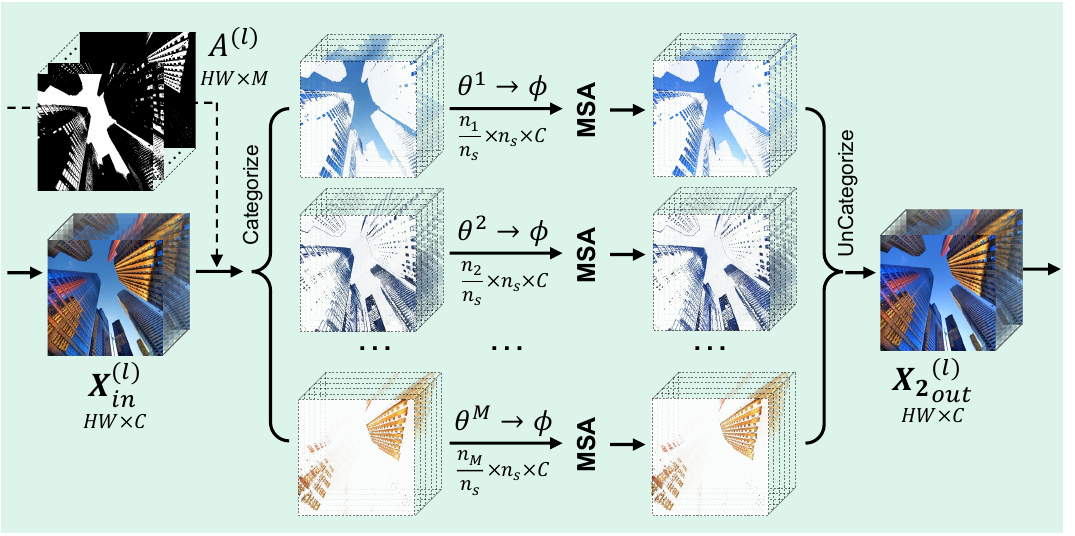}
    \caption{Adaptive Category-based Multi-head Self-Attention}
    \label{fig:acmsa}
  \end{subfigure}
  \caption{The proposed (a) Token Dictionary Cross-Attention (TDCA) and (b) Adaptive Category-based Multi-head Self-Attention (AC-MSA). In \cref{fig:acmsa}, we omit the details of dividing categories $\theta$ into sub-categories $\phi$ for simplicity and better understanding. More details of TDCA and AC-MSA can be found in \cref{sec:TDCA} and \cref{sec:AC-MSA}.
  }
  \label{fig:overall architecture of Transformer Layer}
\end{figure*}

\subsection{Token Dictionary Cross-Attention}
\label{sec:TDCA}
In this subsection, we introduce the details of our proposed token dictionary cross-attention block.
%

In comparison to the existing multi-head self-attention (MSA), which generates query, key, and value tokens by the input feature itself.
We aim to introduce an extra dictionary $\bm{D}\in \mathbb{R}^{M\times d}$, which is initialized as network parameters, to summarize external priors during the training phase.
We use the learned token dictionary $\bm{D}$ 
to generate the Key dictionary $\bm{K}_D$ and the Value dictionary $\bm{V}_D$ and use the input feature $\bm{X}\in \mathbb{R}^{N\times d}$ to generate Query tokens:
\begin{equation}
    \label{eq:KD_VD}
    \bm{Q}_X = \bm{X} \bm{W}^Q, \quad  \bm{K}_D = \bm{D} \bm{W}^K, \quad  \bm{V}_D = \bm{D} \bm{W}^V,
\end{equation}
where $W^Q\in\mathbb{R}^{d\times d/r}$, $W^K\in\mathbb{R}^{d\times d/r}$ and $W^V\in\mathbb{R}^{d\times d}$ are linear transforms for query tokens, key dictionary tokens and value dictionary tokens, respectively.
We set $M \ll N$ to maintain a low computational cost.
Meanwhile, the feature dimensions of query tokens and key dictionary tokens are reduced to $1/r$ to decrease model size and complexity, where $r$ is the reduction ratio.
Then, we use the key dictionary and the value dictionary to enhance query tokens via cross-attention calculation:
\begin{equation}
    \label{eq:TDCA}
    \begin{split}
        & \bm{A} = \operatorname{SoftMax}(\operatorname{Sim_{cos}}(\bm{Q}_X, \bm{K}_D) / \tau),\\
        & \operatorname{TDCA}(\bm{Q}_X, \bm{K}_D, \bm{V}_D)= \bm{A} \cdot \bm{V}_D.
    \end{split}
\end{equation}
In \cref{eq:TDCA}, $\tau$ is a learnable parameter for adjusting the range of similarity value;
$\operatorname{Sim_{cos}}(\cdot,\cdot)$ represents calculating cosine similarity between two tokens, and $\bm{S} = \operatorname{Sim_{cos}}(\bm{Q}_X, \bm{K}_D) \in \mathbb{R}^{N\times M}$ is the similarity map between query image tokens and the key dictionary tokens.
We use the normalized cosine distance instead of the dot product operation in MSA because we want each token in the dictionary to have an equal opportunity to be selected, and the similar magnitude normalization operation is commonly used in previous dictionary learning works.
Then we use a SoftMax function to transform the similarity map $\bm{S}$ to attention map $\bm{A}$ for subsequent calculations.

The above TDCA operation first selects similar tokens in key dictionary and obtains the attention map, which is similar to the sparse representation process in \cref{eq:sr} to obtain representation coefficients;
then TDCA utilizes the similarity values to combine the corresponding tokens in value dictionary, which is the same as reconstructing HR patch with HR dictionary atoms and representation coefficients.
By this way, our TDCA is able to embed the external prior into the learned dictionary to enhance the input image feature.
We will validate the effectiveness of using token dictionary in our ablation study in \cref{sec:ablation}.

\begin{figure*}
    \centering
    \includegraphics[width=\textwidth]{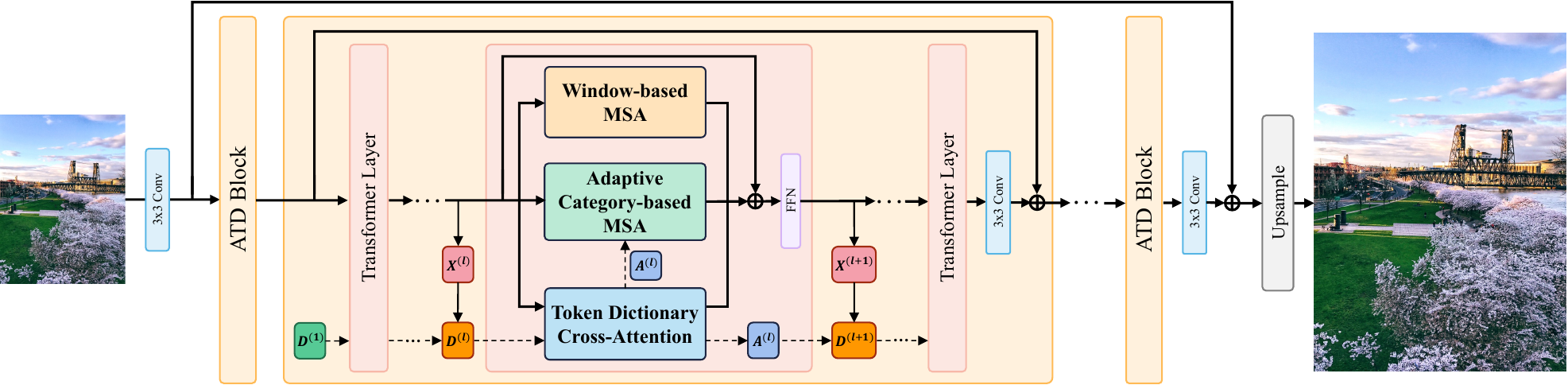}

    
    \captionsetup{font={small}}
    \caption{The overall architecture of the proposed ATD network. Token dictionary cross-attention (\cref{fig:tdca}), adaptive category-based MSA (\cref{fig:acmsa}), and window-based MSA~\cite{liu2021swin} form the main structure of the transformer layer. 
    Each ATD block contains several transformer layers and an initial token dictionary $\bm{D}^{(1)}$. The token dictionary is recurrently adapted via the adaptive dictionary refinement operation.
    }
    \label{fig:overall architecture of ATD}
\end{figure*}

\subsection{Adaptive Dictionary Refinement}
\label{sec:ADR}
In the previous subsection, we have presented how to incorporate extra token dictionary to supply external prior for super-resolution transformer.
Since the image features in each layer are projected to different feature spaces by Multi-Layer Perceptrons (MLPs), we need to learn different Token Dictionary for each layer to provide external prior in each specific feature space.
This will result in a large number of additional parameters.
In this subsection, we introduce an adaptive refining strategy that refines the token dictionary of the previous layer based on the similarity map and updated features in a reversed form of attention.

To introduce the proposed adaptive refining strategy, we set up the layer index $(l)$ for the input features and token dictionary, i.e. $\bm{X}^{(l)}$ and $\bm{D}^{(l)}$ denote input feature and token dictionary of the $l$-th layer, respectively.
We only establish a token dictionary for the initial layer $\bm{D}^{(1)}$ as network parameter discussed in \cref{sec:TDCA} to incorporate external prior knowledge.
In the following layers, each dictionary $\bm{D}^{(l)}$ is refined based on $\bm{D}^{(l-1)}$ from the previous layer.
For each token in the dictionary $\{\bm{d}_i^{(l)}\}_{i=1,\dots,M}$ of the $l$-th layer, we select the corresponding similar tokens in the enhanced feature $\bm{X}^{(l+1)}$, i.e., the output of the $l$-th layer to refine it.
To be more specific, we denote $\bm{a}_i^{(l)}$ as the $i$-th column of attention map $\bm{A}^{(l)}$, which contains the attention weight between $\bm{d}_i^{(l)}$ and all the $N$ query tokens $\bm{X}^{(l)}$.
Therefore, based on each $\bm{a}_i^{(l)}$, we can select the corresponding enhanced tokens $\bm{X}^{(l+1)}$ to reconstruct the new token dictionary element $\bm{d}_i^{(l)}$ and combine them to form $\bm{D}^{(l)}$:
\vspace{-1mm}
\begin{equation}
\vspace{-1mm}
    \label{eq:adr}
    \begin{split}
    & \hat{\bm{D}}^{(l)} = \operatorname{SoftMax}({\operatorname{Norm}(\bm{A}^{(l)}}^T))\bm{X}^{(l+1)}, \\
    & \bm{D}^{(l+1)}     = \sigma \hat{\bm{D}}^{(l)} + (1-\sigma) \bm{D}^{(l)}, 
    \end{split}
\end{equation}
where $\operatorname{Norm}$ is normalization layer to adjust the range of attention map. This refinement can also be perceived as a reverse form of attention in \cref{eq:TDCA}, summarizing the information of updated feature into token dictionary. Then, based on a learnable parameter $\sigma$, we adaptively combine $\hat{\bm{D}}^{(l-1)}$ and $\bm{D}^{(l-1)}$ to obtain $\bm{D}^{(l)}$. In this way, the refined token dictionary is able to integrate both external prior and specific internal prior of the input image.

Due to the linear complexity of the proposed TDCA with the number of image tokens, we do not need to divide the image into windows and $\bm{X}^{(l)}$ represents all image tokens.
Starting from the initial token dictionary $\bm{D}^{(1)}$, which introduces external prior into the network, our adaptive refinement strategy gradually selects relevant tokens from the entire image to refine the dictionary.
The refined dictionary could cross the boundary of self-attention window to summarize the typical local structures of the whole image and consequently improve image feature with global information.
Furthermore, the class information is implicitly embedded in the refined token dictionary.
The attention map $\bm{A}$ contains the similarity relation between the feature and the token dictionary, which is similar to the image classification task to some extent.
The higher similarity between the pixel $x_j$ and a token dictionary atom $\bm{d}_i$ represents the higher probability that $x_j$ belongs to the class of $\bm{d}_i$.
In the next subsection, we utilize this class information to adaptively partition the input feature and propose a category-based self-attention mechanism to achieve non-local attention while keeping an affordable computational cost.

\subsection{Adaptive Category-based Attention}
\label{sec:AC-MSA}

Due to the quadratic computational complexity of self-attention, most of the existing methods, such as Swin Transformer~\cite{liu2021swin}, have to divide the input feature into rectangular windows before performing attention.
Such a window-based attention calculation severely limits the scope of receptive field.
Furthermore, this content-independent partition strategy could lead to unrelated tokens being grouped into the same window, potentially affecting the accuracy of the attention map.

To make better use of self-attention, adaptive feature partitioning could be a more appropriate choice.
Thanks to the attention map between input feature and token dictionary obtained by TDCA, which implicitly incorporates the class information of each pixel, we are able to categorize the input feature.
We classify each pixel into various categories $\bm{\theta}^1, \bm{\theta}^2, \cdots, \bm{\theta}^M$ based on which dictionary token is most similar to:
\begin{equation}
\label{eq:categorize}
    \bm{\theta}^i = \{ x_j | \operatorname{arg}\operatorname{max}_k (\bm{A}_{jk}) = i \},
\end{equation}
where $\bm{A} \in \mathbb{R}^{N\times M}$ is the attention map obtained by \cref{eq:TDCA}.
The pixel $x_j$ will be classified into $\bm{\theta}^i$ if $\bm{A}_{ji}$ is the highest among $\bm{A}_{j1}, \bm{A}_{j2}, \cdots, \bm{A}_{jM}$, which indicates that $x_j$ is more likely to be of the same class as the $i$-th token $\bm{d}_i$ in the dictionary.
Therefore, each category can be perceived as an irregularly shaped window that contains tokens of the same class.
An example of categorization visualization is presented in \cref{fig:visualization of acmsa}.
However, the number of tokens in each category may differ, which results in low parallelism efficiency and significant computational burden.
To address the issue of unbalanced categorization, we refer to \cite{Mei_2021_nlsa} to further divide the categories $\bm{\theta}$ into sub-categories $\bm{\phi}$:
\begin{equation}
\label{eq:subcategorize}
    \begin{split}
        \bm{\phi} &= \left[ \bm{\theta}_1^1, \bm{\theta}_2^1, \cdots, \bm{\theta}_{n_1}^1, \cdots, \bm{\theta}_{n_M}^M \right], \\
        \bm{\phi}^{j} &= \left[ \bm{\phi}_{j * n_s + 1}, \bm{\phi}_{j * n_s + 2}, \cdots, \bm{\phi}_{(j+1) * n_s} \right], 
    \end{split}
\end{equation}
where the category $\bm{\theta}^{i}$ contains $n_i$ tokens. 
Each category is flattened and concatenated to form $\bm{\phi}$, then divided into sub-categories $[\bm{\phi}^{1}, \bm{\phi}^{2}, \cdots, \bm{\phi}^{j}, \cdots]$.
After division, all subcategories have the same fixed size $n_{s}$, improving parallelism efficiency.
The illustrations are presented in \cref{fig:visualization of categorize} and \cref{fig:more visualization ac-msa}.
In general, the procedure of AC-MSA can be formulated as:
\begin{equation}
\label{eq:AC-MSA}
    \begin{split}
        \{ \bm{\phi}^{j}\} &= \operatorname{Categorize}(\bm{X}_{in}) , \\
        \hat{\bm{\phi}}^{j} &= \operatorname{MSA}(\bm{\phi}^{j} \bm{W}^{Q}, \bm{\phi}^{j} \bm{W}^{K}, \bm{\phi}^{j} \bm{W}^{V}), \\
        \bm{X}_{out} &= \operatorname{UnCategorize} (\{ \hat{\bm{\phi}}^{j} \}),
    \end{split}
\end{equation}
where the $\operatorname{Categorize}$ operation is the combination of \cref{eq:categorize} and \cref{eq:subcategorize} that divides input feature into categories and further into sub-categories. 
Then, we could view each sub-category as an attention group and perform multi-head self-attention within each group.
Finally, we use the $\operatorname{UnCategorize}$ operation (inversed $\operatorname{Categorize}$ operation) to put each pixel back to its original position on the feature map to form $\bm{X}_{out}$.

Although the sub-category division limits the size of each attention group, it only slightly affects the receptive field.
This is due to the random shuffle that occurred during the sort operation for division in \cref{eq:subcategorize}.
Therefore, each sub-category can be viewed as a random sample from the category.
Tokens in a certain sub-category could still be spread over the entire feature map, maintaining a global receptive field.
In general, the proposed AC-MSA classifies similar features into the same category and performs attention within each category, 
breaking through the limitation of window partitioning and establishing global connections between similar features.
We will conduct ablation studies and provide visualization of the categorization results in later sections to quantitatively and qualitatively verify the effectiveness of AC-MSA.


 \subsection{The Overall Network Architecture}
Having our proposed token dictionary cross-attention (TDCA), adaptive dictionary refinement (ADR) strategy, and adaptive category-based multi-head self-attention (AC-MSA), we are able to establish our Adaptive Token Dictionary (ATD) network for image super-resolution.
As shown in \cref{fig:overall architecture of ATD}, given an input low-resolution image, we first utilize a 3$\times$3 convolution layer to extract shallow features.
The shallow features are then fed into a series of ATD blocks,
where each ATD block contains several ATD transformer layers.
We combine token dictionary cross-attention, adaptive category-based multi-head self-attention, and the commonly used shift window-based multi-head self-attention~\cite{liang2021swinir, liu2021swin} to form the transformer layer.
These three attention modules work in parallel to take advantage of external, global, and local features of the input feature.
Then, the features are combined by a summation operation.
In addition to the attention module, our transformer layer also utilizes the LayerNorm and FFN layers, which have been commonly utilized in other transformer-based architectures.
Moreover, the token dictionary begins with the learnable parameters within each ATD block. 
It takes part in the token dictionary cross-attention of each transformer layer, and we utilize the adaptive dictionary refinement strategy to adapt the dictionary to the input feature for the next layer.
After the ATD blocks, we utilize an extra convolution layer followed by a pixel shuffle operation to generate the final HR estimation.

\section{Experiments}

\subsection{Experimental Settings}
We propose the ATD model that employs a sequence of ATD blocks as its backbone.
There are six ATD blocks in total, each comprising six transformer layers with a channel number of 210.
We establish 128 tokens for our external token dictionary $\bm{D}^{(1)}$ in each ATD-block and use a reduction rate $r=10.5$ to decrease the channel number to 20 for similarity calculation.
Each dictionary is randomly initialized as a tensor of shape $[128, 210]$ in normal distribution.
For the adaptive category-based attention branch, the sub-categories size $n_s$ is set to 128.
Furthermore, we establish ATD-light as a lightweight version of ATD with 48 feature dimensions and 4 ATD blocks for lightweight SR task.
The number of tokens in each dictionary is reduced to 64, and we also adjust the reduction rate to $r=6$ to maintain eight dimensions during the similarity calculation.
Details of training procedure can be found in the supplementary material.

\begin{table}
\vspace{-0mm}
 \centering
 \caption{Ablation study on the effects of each component. Detailed experimental settings can be found in our Ablation study section.}
 \label{tab:ablation on td adr}
    \captionsetup{font={small}}
    \footnotesize
    \setlength{\tabcolsep}{5.6pt}
    \begin{tabular}{|ccc|cc|cc|}
        \hline
        \multirow{2}{*}{\textbf{TDCA}} & \multirow{2}{*}{\textbf{ADR}} & \multirow{2}{*}{\textbf{AC-MSA}} & \multicolumn{2}{c|}{\textbf{Urban100}} & \multicolumn{2}{c|}{\textbf{Manga109}} \\

        & & & PSNR & SSIM & PSNR & SSIM   \\
        \hline \hline
                &         &         & 26.25 & 0.7907 & 30.66 & 0.9113 \\
        $\surd$ &         &         & 26.32 & 0.7929 & 30.76 & 0.9118 \\
        $\surd$ & $\surd$ &         & 26.36 & 0.7931 & 30.79 & 0.9123 \\
        $\surd$ & $\surd$ & $\surd$ & 26.51 & 0.7975 & 30.98 & 0.9144 \\
        \hline
    \end{tabular}
\end{table}

\begin{table}
 \centering
 \caption{Ablation study on different designs of category-based attention. CA denotes category-based attention. }
 \label{tab:ablation on category}
    \captionsetup{font={small}}
    \footnotesize
    \setlength{\tabcolsep}{4.6pt}
    \begin{tabular}{|l|cc|cc|cc|}
        \hline
        \multirow{2}{*}{\textbf{Model}} & \multicolumn{2}{c|}{\textbf{Set5}} & \multicolumn{2}{c|}{\textbf{Urban100}} & \multicolumn{2}{c|}{\textbf{Manga109}} \\

         & PSNR & SSIM & PSNR & SSIM & PSNR & SSIM   \\
        \hline \hline
        w/o CA                 & 32.30 & 0.8957 & 26.25 & 0.7907 & 30.66 & 0.9113 \\
        \textbf{random} CA     & 32.38 & 0.8962 & 26.46 & 0.7955 & 30.92 & 0.9139 \\
        \textbf{adaptive} CA   & 32.46 & 0.8973 & 26.51 & 0.7975 & 30.98 & 0.9144 \\
        \hline
    \end{tabular}
    
\end{table}

\begin{table}
 \centering
 \caption{Ablation study on sub-category size $n_s$ and dictionary size $M$. The best results are \textbf{highlighted}.}
 \label{tab:ablation on subcategory size}
    \captionsetup{font={small}}
    \footnotesize
    \setlength{\tabcolsep}{2pt}
    
    \begin{tabular}{|c|cc|cc||c|cc|cc|}
        \hline
        
        \multirow{2}{*}{\textbf{$n_s$}} & \multicolumn{2}{c|}{\textbf{Urban100}} & \multicolumn{2}{c||}{\textbf{Manga109}} & \multirow{2}{*}{\textbf{$M$}} & \multicolumn{2}{c|}{\textbf{Urban100}} & \multicolumn{2}{c|}{\textbf{Manga109}} \\
        & PSNR & SSIM & PSNR & SSIM & & PSNR & SSIM & PSNR & SSIM \\
        \hline \hline
        0      & 26.36 & 0.7931 & 30.79 & 0.9123 & 16     & 26.45 & 0.7950 & 30.90 & 0.9137 \\
        64     & 26.44 & 0.7948 & 30.90 & 0.9131 & 32     & 26.49 & 0.7965 & 30.97 & 0.9142 \\
        128    & 26.51 & 0.7975 & 30.98 & 0.9144 & 64     & \textbf{26.51} & \textbf{0.7975} & \textbf{30.98} & \textbf{0.9144} \\
        192    & \textbf{26.55} & \textbf{0.7984} & \textbf{31.01} & \textbf{0.9150} & 96     & \textbf{26.51} & 0.7970 & 30.95 & 0.9141 \\
        
        \hline
    \end{tabular}
    
\end{table}

\begin{table*}
\captionsetup{font={small}}
\scriptsize
  
\caption{Quantitative comparison (PSNR/SSIM) with state-of-the-art methods on \textbf{classical SR} task. The best and second best results are colored with \sotaa{red} and \sotab{blue}. }
\vspace{-3mm}
\label{tab: results image sr1}
  \begin{center}
      
  \begin{tabular}{|p{1.8cm}|c|c|cc|cc|cc|cc|cc|}
    \hline
    \multirow{2}{*}{\textbf{Method}} & \multirow{2}{*}{\textbf{Scale}} & \multirow{2}{*}{\textbf{Params}} & \multicolumn{2}{c|}{\textbf{Set5}} & \multicolumn{2}{c|}{\textbf{Set14}} & \multicolumn{2}{c|}{\textbf{BSD100}} & \multicolumn{2}{c|}{\textbf{Urban100}} & \multicolumn{2}{c|}{\textbf{Manga109}} \\

    & & & PSNR & SSIM & PSNR & SSIM & PSNR & SSIM & PSNR & SSIM & PSNR & SSIM   \\

    \hline

    EDSR~\cite{lim2017edsr}         & $\times$2 & 42.6M & 38.11 & 0.9602 & 33.92 & 0.9195 & 32.32 & 0.9013 & 32.93 & 0.9351 & 39.10 & 0.9773 \\
    RCAN~\cite{zhang2018rcan}       & $\times$2 & 15.4M & 38.27 & 0.9614 & 34.12 & 0.9216 & 32.41 & 0.9027 & 33.34 & 0.9384 & 39.44 & 0.9786 \\
    SAN~\cite{Dai_2020_san}         & $\times$2 & 15.7M & 38.31 & 0.9620 & 34.07 & 0.9213 & 32.42 & 0.9028 & 33.10 & 0.9370 & 39.32 & 0.9792 \\
    HAN~\cite{Niu_2020_han}         & $\times$2 & 63.6M & 38.27 & 0.9614 & 34.16 & 0.9217 & 32.41 & 0.9027 & 33.35 & 0.9385 & 39.46 & 0.9785 \\
    IPT~\cite{Chen_2020_ipt}        & $\times$2 & 115M  & 38.37 & - & 34.43 & - & 32.48 & - & 33.76 & - & - & - \\
    SwinIR~\cite{liang2021swinir}   & $\times$2 & 11.8M & 38.42 & 0.9623 & 34.46 & 0.9250 & 32.53 & 0.9041 & 33.81 & 0.9433 & 39.92 & 0.9797 \\
    EDT~\cite{li2021efficient}      & $\times$2 & 11.5M & 38.45 & 0.9624 & 34.57 & 0.9258 & 32.52 & 0.9041 & 33.80 & 0.9425 & 39.93 & 0.9800 \\
    CAT-A~\cite{chen2022cross}      & $\times$2 & 16.5M & 38.51 & 0.9626 & 34.78 & 0.9265 & 32.59 & 0.9047 & 34.26 & 0.9440 & 40.10 & 0.9805 \\
    ART~\cite{zhang2023accurate}    & $\times$2 & 16.4M & 38.56 & 0.9629 & 34.59 & 0.9267 & 32.58 & 0.9048 & 34.30 & 0.9452 & 40.24 & 0.9808 \\
    HAT~\cite{chen2023activating}   & $\times$2 & 20.6M & \sotaa{38.63} & \sotaa{0.9630} & \sotab{34.86} & \sotab{0.9274} & \sotab{32.62} & \sotab{0.9053} & \sotab{34.45} & \sotab{0.9466} & \sotab{40.26} & \sotab{0.9809} \\
    
    \rowcolor{Gray}
    \textbf{ATD} (ours)              & $\times$2 & 20.1M & \sotab{38.61} & \sotab{0.9629} & \sotaa{34.95} & \sotaa{0.9276} & \sotaa{32.65} & \sotaa{0.9056} & \sotaa{34.70} & \sotaa{0.9476} & \sotaa{40.37} & \sotaa{0.9810} \\

    \hline \hline

    EDSR~\cite{lim2017edsr}         & $\times$3 & 43.0M & 34.65 & 0.9280 & 30.52 & 0.8462 & 29.25 & 0.8093 & 28.80 & 0.8653 & 34.17 & 0.9476 \\
    RCAN~\cite{zhang2018rcan}       & $\times$3 & 15.6M & 34.74 & 0.9299 & 30.65 & 0.8482 & 29.32 & 0.8111 & 29.09 & 0.8702 & 34.44 & 0.9499 \\
    SAN~\cite{Dai_2020_san}         & $\times$3 & 15.9M & 34.75 & 0.9300 & 30.59 & 0.8476 & 29.33 & 0.8112 & 28.93 & 0.8671 & 34.30 & 0.9494 \\
    HAN~\cite{Niu_2020_han}         & $\times$3 & 64.2M & 34.75 & 0.9299 & 30.67 & 0.8483 & 29.32 & 0.8110 & 29.10 & 0.8705 & 34.48 & 0.9500 \\
    IPT~\cite{Chen_2020_ipt}        & $\times$3 & 116M & 34.81 & -      & 30.85 & -      & 29.38 & -      & 29.49 & -      & -     & -      \\
    SwinIR~\cite{liang2021swinir}   & $\times$3 & 11.9M & 34.97 & 0.9318 & 30.93 & 0.8534 & 29.46 & 0.8145 & 29.75 & 0.8826 & 35.12 & 0.9537 \\
    EDT~\cite{li2021efficient}      & $\times$3 & 11.6M & 34.97 & 0.9316 & 30.89 & 0.8527 & 29.44 & 0.8142 & 29.72 & 0.8814 & 35.13 & 0.9534 \\
    CAT-A~\cite{chen2022cross}      & $\times$3 & 16.6M & 35.06 & 0.9326 & 31.04 & 0.8538 & 29.52 & 0.8160 & 30.12 & 0.8862 & 35.38 & 0.9546 \\
    ART~\cite{zhang2023accurate}    & $\times$3 & 16.6M & \sotab{35.07} & 0.9325 & 31.02 & 0.8541 & 29.51 & 0.8159 & 30.10 & 0.8871 & 35.39 & 0.9548 \\
    HAT~\cite{chen2023activating}   & $\times$3 & 20.8M & \sotab{35.07} & \sotab{0.9329} & \sotab{31.08} & \sotab{0.8555} & \sotab{29.54} & \sotab{0.8167} & \sotab{30.23} & \sotab{0.8896} & \sotab{35.53} & \sotab{0.9552} \\

    \rowcolor{Gray}
    \textbf{ATD} (ours)             & $\times$3 & 20.3M & \sotaa{35.11} & \sotaa{0.9330} & \sotaa{31.13} & \sotaa{0.8556} & \sotaa{29.57} & \sotaa{0.8176} & \sotaa{30.46} & \sotaa{0.8917} & \sotaa{35.63} & \sotaa{0.9558} \\

    \hline \hline
    
    EDSR~\cite{lim2017edsr}         & $\times$4 & 43.0M & 32.46 & 0.8968 & 28.80 & 0.7876 & 27.71 & 0.7420 & 26.64 & 0.8033 & 31.02 & 0.9148 \\
    RCAN~\cite{zhang2018rcan}       & $\times$4 & 15.6M & 32.63 & 0.9002 & 28.87 & 0.7889 & 27.77 & 0.7436 & 26.82 & 0.8087 & 31.22 & 0.9173 \\
    SAN~\cite{Dai_2020_san}         & $\times$4 & 15.9M & 32.64 & 0.9003 & 28.92 & 0.7888 & 27.78 & 0.7436 & 26.79 & 0.8068 & 31.18 & 0.9169 \\
    HAN~\cite{Niu_2020_han}         & $\times$4 & 64.2M & 32.64 & 0.9002 & 28.90 & 0.7890 & 27.80 & 0.7442 & 26.85 & 0.8094 & 31.42 & 0.9177 \\
    IPT~\cite{Chen_2020_ipt}        & $\times$4 & 116M  & 32.64 & -      & 29.01 & -      & 27.82 & -      & 27.26 & -      & -     & -      \\   
    SwinIR~\cite{liang2021swinir}   & $\times$4 & 11.9M & 32.92 & 0.9044 & 29.09 & 0.7950 & 27.92 & 0.7489 & 27.45 & 0.8254 & 32.03 & 0.9260 \\
    EDT~\cite{li2021efficient}      & $\times$4 & 11.6M & 32.82 & 0.9031 & 29.09 & 0.7939 & 27.91 & 0.7483 & 27.46 & 0.8246 & 32.05 & 0.9254 \\
    CAT-A~\cite{chen2022cross}      & $\times$4 & 16.6M & \sotab{33.08} & 0.9052 & 29.18 & 0.7960 & 27.99 & 0.7510 & 27.89 & 0.8339 & 32.39 & 0.9285 \\
    ART~\cite{zhang2023accurate}    & $\times$4 & 16.6M & 33.04 & 0.9051 & 29.16 & 0.7958 & 27.97 & 0.7510 & 27.77 & 0.8321 & 32.31 & 0.9283 \\
    HAT~\cite{chen2023activating}   & $\times$4 & 20.8M & 33.04 & \sotab{0.9056} & \sotab{29.23} & \sotab{0.7973} & \sotab{28.00} & \sotab{0.7517} & \sotab{27.97} & \sotab{0.8368} & \sotab{32.48} & \sotab{0.9292} \\
    
    \rowcolor{Gray}
    \textbf{ATD} (ours)             & $\times$4 & 20.3M & \sotaa{33.10} & \sotaa{0.9058} & \sotaa{29.24} & \sotaa{0.7974} & \sotaa{28.01} & \sotaa{0.7526} & \sotaa{28.17} & \sotaa{0.8404} & \sotaa{32.62} & \sotaa{0.9306} \\
    \hline
  \end{tabular}
  \end{center}
\end{table*}

\begin{table*}
\captionsetup{font={small}}
\scriptsize
  
\caption{Quantitative comparison (PSNR/SSIM) with state-of-the-art methods on \textbf{lightweight SR} task. The best and second best results are colored with \sotaa{red} and \sotab{blue}. }
\vspace{-3mm}
\label{tab: results image sr2}
  \begin{center}
      
  \begin{tabular}{|p{1.8cm}|c|c|cc|cc|cc|cc|cc|}
    \hline
    \multirow{2}{*}{\textbf{Method}} & \multirow{2}{*}{\textbf{Scale}} & \multirow{2}{*}{\textbf{Params}} & \multicolumn{2}{c|}{\textbf{Set5}} & \multicolumn{2}{c|}{\textbf{Set14}} & \multicolumn{2}{c|}{\textbf{BSD100}} & \multicolumn{2}{c|}{\textbf{Urban100}} & \multicolumn{2}{c|}{\textbf{Manga109}} \\

    & & & PSNR & SSIM & PSNR & SSIM & PSNR & SSIM & PSNR & SSIM & PSNR & SSIM   \\

    \hline \hline
    CARN~\cite{Ahn_2018_carn}               & $\times$2 & 1,592K & 37.76 & 0.9590 & 33.52 & 0.9166 & 32.09 & 0.8978 & 31.92 & 0.9256 & 38.36 & 0.9765 \\
    IMDN~\cite{Hui_2019_imdn}               & $\times$2 & 694K   & 38.00 & 0.9605 & 33.63 & 0.9177 & 32.19 & 0.8996 & 32.17 & 0.9283 & 38.88 & 0.9774 \\
    LAPAR-A~\cite{Li_2020_lapar}            & $\times$2 & 548K   & 38.01 & 0.9605 & 33.62 & 0.9183 & 32.19 & 0.8999 & 32.10 & 0.9283 & 38.67 & 0.9772 \\
    LatticeNet~\cite{Luo_2020_latticenet}   & $\times$2 & 756K   & 38.15 & 0.9610 & 33.78 & 0.9193 & 32.25 & 0.9005 & 32.43 & 0.9302 & -     & -      \\
    SwinIR-light~\cite{liang2021swinir}     & $\times$2 & 910K   & 38.14 & 0.9611 & 33.86 & 0.9206 & 32.31 & 0.9012 & 32.76 & 0.9340 & 39.12 & 0.9783 \\
    ELAN~\cite{zhang2022efficient}          & $\times$2 & 582K   & 38.17 & 0.9611 & 33.94 & 0.9207 & 32.30 & 0.9012 & 32.76 & 0.9340 & 39.11 & 0.9782 \\
    SwinIR-NG~\cite{Choi_2022_swinirng}     & $\times$2 & 1181K  & 38.17 & 0.9612 & 33.94 & 0.9205 & 32.31 & 0.9013 & 32.78 & 0.9340 & 39.20 & 0.9781 \\
    OmniSR~\cite{omni_sr}                   & $\times$2 & 772K   & \sotab{38.22} & \sotab{0.9613} & \sotab{33.98} & \sotab{0.9210} & \sotab{32.36} & \sotab{0.9020} & \sotab{33.05} & \sotab{0.9363} & \sotab{39.28} & \sotab{0.9784} \\
    
    \rowcolor{Gray}
    \textbf{ATD-light} (Ours)      & $\times$2 & 753K   & \sotaa{38.28} & \sotaa{0.9616} & \sotaa{34.11} & \sotaa{0.9217} & \sotaa{32.39} & \sotaa{0.9023} & \sotaa{33.27} & \sotaa{0.9376} & \sotaa{39.51} & \sotaa{0.9789} \\

    \hline \hline

    CARN~\cite{Ahn_2018_carn}               & $\times$3 & 1,592K & 34.29 & 0.9255 & 30.29 & 0.8407 & 29.06 & 0.8034 & 28.06 & 0.8493 & 33.50 & 0.9440 \\
    IMDN~\cite{Hui_2019_imdn}               & $\times$3 & 703K   & 34.36 & 0.9270 & 30.32 & 0.8417 & 29.09 & 0.8046 & 28.17 & 0.8519 & 33.61 & 0.9445 \\
    LAPAR-A~\cite{Li_2020_lapar}            & $\times$3 & 544K   & 34.36 & 0.9267 & 30.34 & 0.8421 & 29.11 & 0.8054 & 28.15 & 0.8523 & 33.51 & 0.9441 \\
    LatticeNet~\cite{Luo_2020_latticenet}   & $\times$3 & 765K   & 34.53 & 0.9281 & 30.39 & 0.8424 & 29.15 & 0.8059 & 28.33 & 0.8538 & -     & -      \\
    SwinIR-light~\cite{liang2021swinir}     & $\times$3 & 918K   & 34.62 & 0.9289 & 30.54 & 0.8463 & 29.20 & 0.8082 & 28.66 & 0.8624 & 33.98 & 0.9478 \\
    ELAN~\cite{zhang2022efficient}          & $\times$3 & 590K   & 34.61 & 0.9288 & 30.55 & 0.8463 & 29.21 & 0.8081 & 28.69 & 0.8624 & 34.00 & 0.9478 \\
    SwinIR-NG~\cite{Choi_2022_swinirng}     & $\times$3 & 1190K  & 34.64 & 0.9293 & \sotab{30.58} & \sotab{0.8471} & 29.24 & 0.8090 & 28.75 & 0.8639 & \sotab{34.22} & \sotab{0.9488} \\
    OmniSR~\cite{omni_sr}                   & $\times$3 & 780K   & \sotaa{34.70} & \sotab{0.9294} & 30.57 & 0.8469 & \sotab{29.28} & \sotab{0.8094} & \sotab{28.84} & \sotab{0.8656} & \sotab{34.22} & 0.9487 \\
    \rowcolor{Gray}
    \textbf{ATD-light} (ours)               & $\times$3 & 760K   & \sotaa{34.70} & \sotaa{0.9300} & \sotaa{30.68} & \sotaa{0.8485} & \sotaa{29.32} & \sotaa{0.8109} & \sotaa{29.16} & \sotaa{0.8710} & \sotaa{34.60} & \sotaa{0.9505} \\

    \hline \hline
    CARN~\cite{Ahn_2018_carn}               & $\times$4 & 1,592K & 32.13 & 0.8937 & 28.60 & 0.7806 & 27.58 & 0.7349 & 26.07 & 0.7837 & 30.47 & 0.9084 \\
    IMDN~\cite{Hui_2019_imdn}               & $\times$4 & 715K   & 32.21 & 0.8948 & 28.58 & 0.7811 & 27.56 & 0.7353 & 26.04 & 0.7838 & 30.45 & 0.9075 \\
    LAPAR-A~\cite{Li_2020_lapar}            & $\times$4 & 659K   & 32.15 & 0.8944 & 28.61 & 0.7818 & 27.61 & 0.7366 & 26.14 & 0.7871 & 30.42 & 0.9074 \\
    LatticeNet~\cite{Luo_2020_latticenet}   & $\times$4 & 777K   & 32.30 & 0.8962 & 28.68 & 0.7830 & 27.62 & 0.7367 & 26.25 & 0.7873 & -     & -      \\
    SwinIR-light~\cite{liang2021swinir}     & $\times$4 & 930K   & 32.44 & 0.8976 & 28.77 & 0.7858 & 27.69 & 0.7406 & 26.47 & 0.7980 & 30.92 & 0.9151 \\
    ELAN~\cite{zhang2022efficient}          & $\times$4 & 582K   & 32.43 & 0.8975 & 28.78 & 0.7858 & 27.69 & 0.7406 & 26.54 & 0.7982 & 30.92 & 0.9150 \\
    SwinIR-NG~\cite{Choi_2022_swinirng}     & $\times$4 & 1201K  & 32.44 & 0.8980 & \sotab{28.83} & \sotab{0.7870} & \sotab{27.73} & \sotab{0.7418} & 26.61 & 0.8010 & \sotab{31.09} & \sotab{0.9161} \\
    OmniSR~\cite{omni_sr}                   & $\times$4 & 792K   & \sotab{32.49} & \sotab{0.8988} & 28.78 & 0.7859 & 27.71 & 0.7415 & \sotab{26.65} & \sotab{0.8018} & 31.02 & 0.9151 \\
    
    \rowcolor{Gray}
    \textbf{ATD-light} (Ours)      & $\times$4 & 769K   & \sotaa{32.62} & \sotaa{0.8997} & \sotaa{28.87} & \sotaa{0.7884} & \sotaa{27.77} & \sotaa{0.7439} & \sotaa{26.97} & \sotaa{0.8107} & \sotaa{31.47} & \sotaa{0.9198} \\
    \hline
  \end{tabular}
  \end{center}
\vspace{-8mm}
\end{table*}

\subsection{Ablation Study}
\label{sec:ablation}

We perform ablation studies on the rescaled ATD-light model and train all models for 250k iterations on the DIV2K~\cite{timofte2017div2k} dataset. We then evaluate them on Set5~\cite{Bevilacqua2012set5}, Urban100~\cite{Huang_2015_Urban100}, and Manga109~\cite{Matsui_2016_Manga109} benchmarks.

\paragraph{Effects of TDCA, ADR, and AC-MSA.}
In order to show the effectiveness of several key design choices in the proposed adaptive token dictionary (ATD) model, we establish four models and compare their ability for image SR.
The first model is the baseline model; we remove the TDCA and AC-MSA branch and only adopt the SW-MSA block to process image features.
To demonstrate the effectiveness of learned token dictionary and token dictionary cross-attention, we present the second model, which directly learns an external token dictionary for each Transformer layer.
In the third model, we employ the adaptive dictionary refinement strategy to tailor the learned token dictionary to the specific input feature.
As shown in \cref{tab:ablation on td adr}, the TDCA branch and the ADR strategy jointly produce 0.11 dB and 0.13 dB improvement on Urban100 and Manga109 datasets respectively.
Furthermore, equipped with adaptive category-based MSA, the final model achieves the best performance of 26.51 / 30.98 dB on the Urban100 / Manga109 benchmark.
These experimental results clearly demonstrate the advantages of TDCA, ADR, and AC-MSA.

\paragraph{Effects of different designs of category-based attention.}
We conduct experiments to explore the effectiveness of the category-based partition strategy.
First, we evaluate the advantages of category-based attention, using a random token dictionary for rough categorization. 
The results in \cref{tab:ablation on category} demonstrate that this random category-based attention still performs better than the baseline.
Then, with the learned adaptive token dictionary, we can perform the categorization procedure more accurately.
The more precise categorization leads to better partition results, resulting in an extra performance gain of 0.05-0.08 dB when using adaptive category-based attention, as opposed to the random one.

\paragraph{Effects of sub-category size $n_s$.}
Increasing the window size  is essential for  window-based attention.
A larger window size provides a wider range of receptive fields, which in turn leads to improved performance.
We carry out experiments to explore the influence of varying sub-category sizes $n_s$ from 0 to 192 on AC-MSA, where 0 represents the removal of the category-based branch, as illustrated in \cref{tab:ablation on subcategory size}.
The model is significantly improved when the value of $n_s$ is raised to 128.
However, when we continue increasing $n_s$, the model performance improves slowly.
This is because AC-MSA has the ability to model long-range dependencies with an appropriate sub-category size.
The larger $n_s$ contributes less to the receptive field and reconstruction accuracy.
To balance performance and computational resource consumption, we set $n_s=128$ for our final model.

\paragraph{Effects of token dictionary size $M$.}
In the token dictionary cross-attention branch, we initialize the token dictionary as M learnable vectors.
We investigate the performance change by gradually increasing the dictionary size from 16 to 96.
As shown in \cref{tab:ablation on subcategory size}, increasing the dictionary size yields an improvement of $0.06 - 0.08$ dB on the evaluation benchmark at first.
However, when $M$ is set to 96, the model even shows performance degradation.
It indicates that the excess of tokens exceeds the modeling capacity of the model and results in unsatisfactory outcomes.


\begin{figure*}
    \centering
    
    \captionsetup{font={small}}
    \vspace{0mm}
    \includegraphics[width=.49\textwidth]{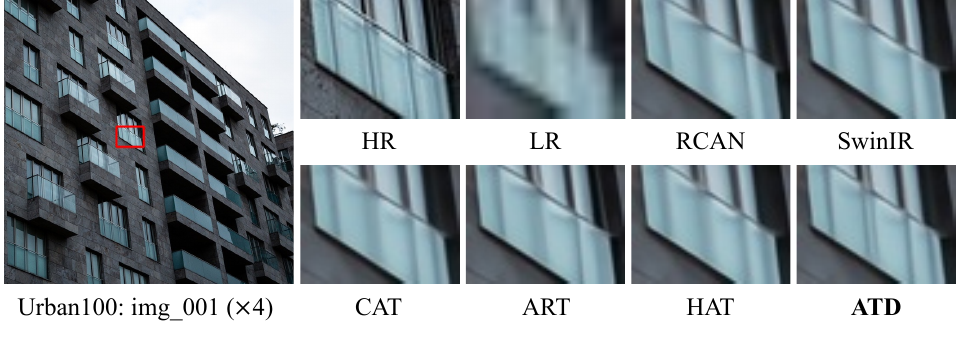}
    \hfill
    \includegraphics[width=.49\textwidth]{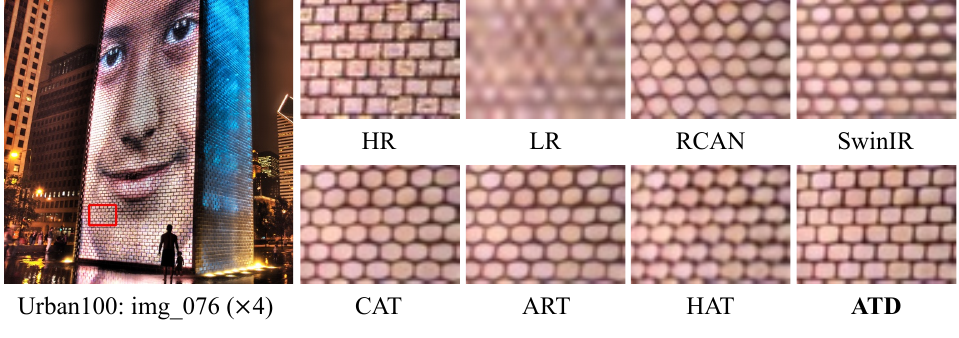}
    
    \vspace{3mm}

    \includegraphics[width=.49\textwidth]{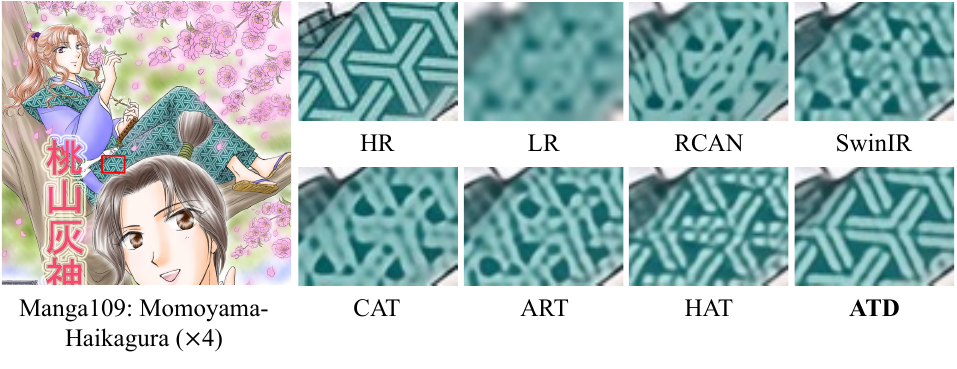}
    \hfill
    \includegraphics[width=.49\textwidth]{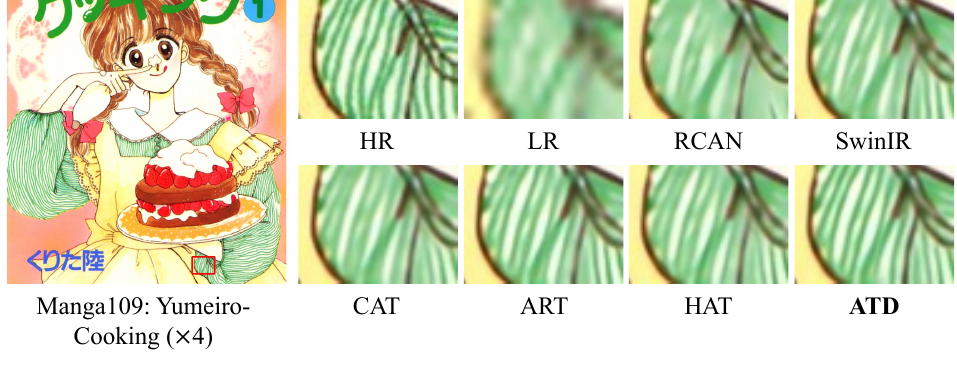}
    
    \caption{Visual comparisons of \textbf{ATD} and other state-of-the-art image super-resolution methods.}
    \label{fig:visualization 1}
    
\end{figure*}

\begin{figure*}
\vspace{3mm}
    \centering
    
    \captionsetup{font={small}}
    \begin{subfigure}{.19\linewidth}
        \centering
        \includegraphics[width=\linewidth]{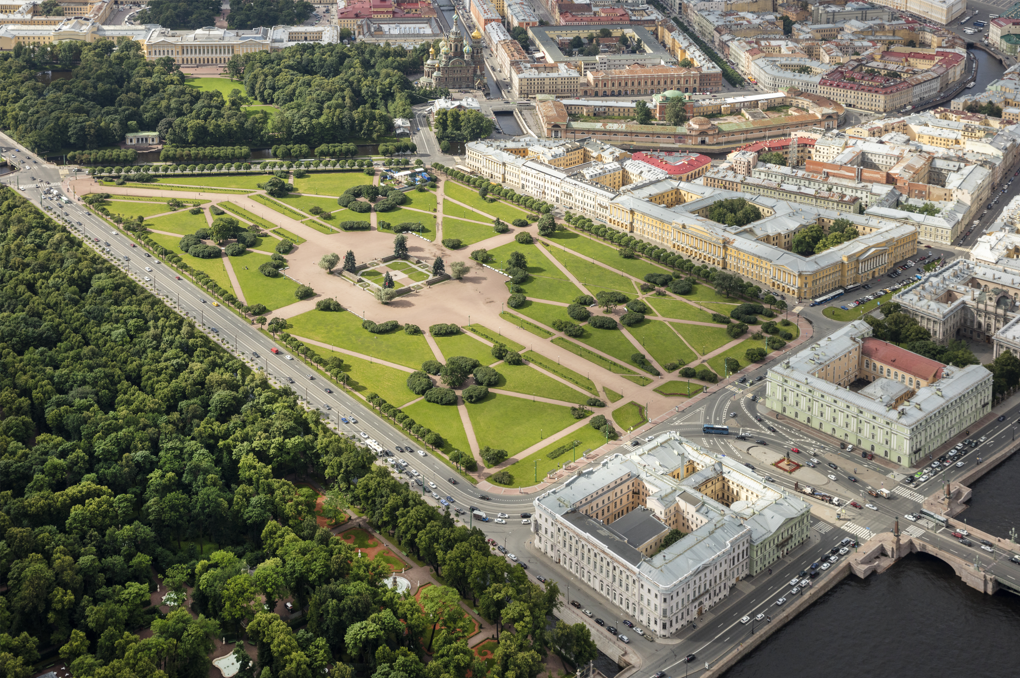}
        \caption{ }
        \label{subfig:vis1}
    \end{subfigure}%
    \hfill
    \begin{subfigure}{.19\linewidth}
        \centering
        \includegraphics[width=\linewidth]{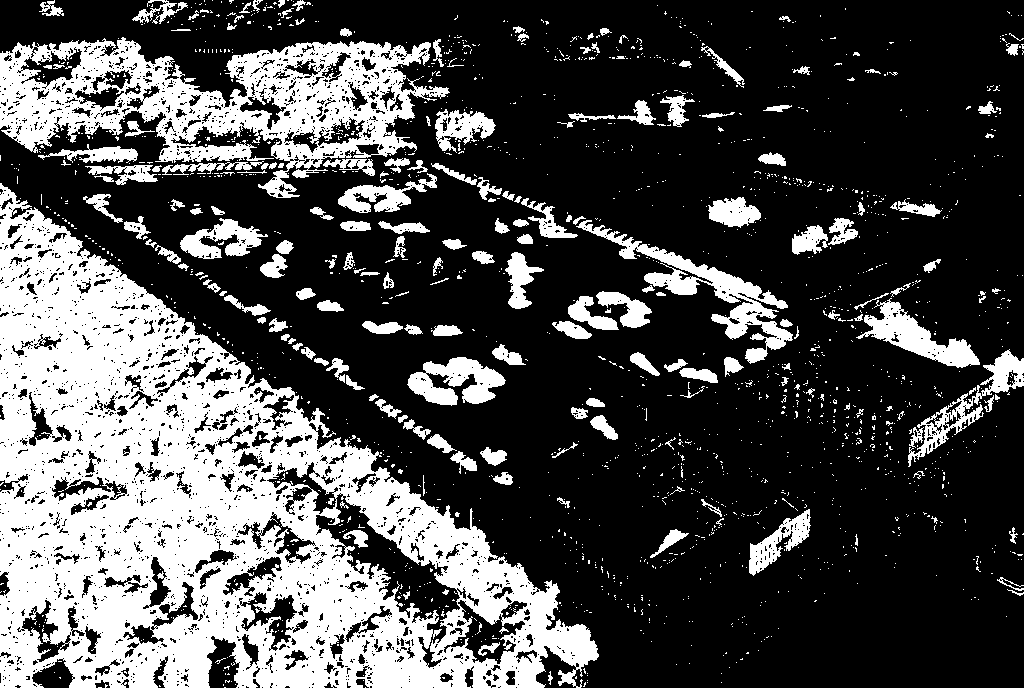}
        \caption{ }
        \label{subfig:vis2}
    \end{subfigure}
    \hfill
    \begin{subfigure}{.19\linewidth}
        \centering
        \includegraphics[width=\linewidth]{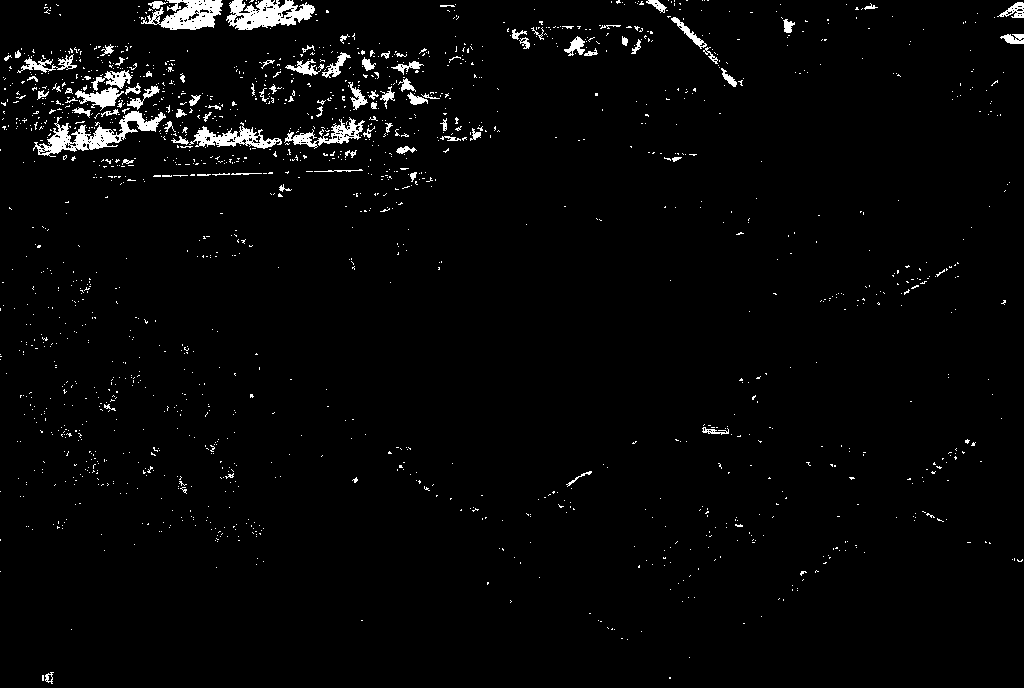}
        \caption{ }
        \label{subfig:vis3}
    \end{subfigure}%
    \hfill
    \begin{subfigure}{.19\linewidth}
        \centering
        \includegraphics[width=\linewidth]{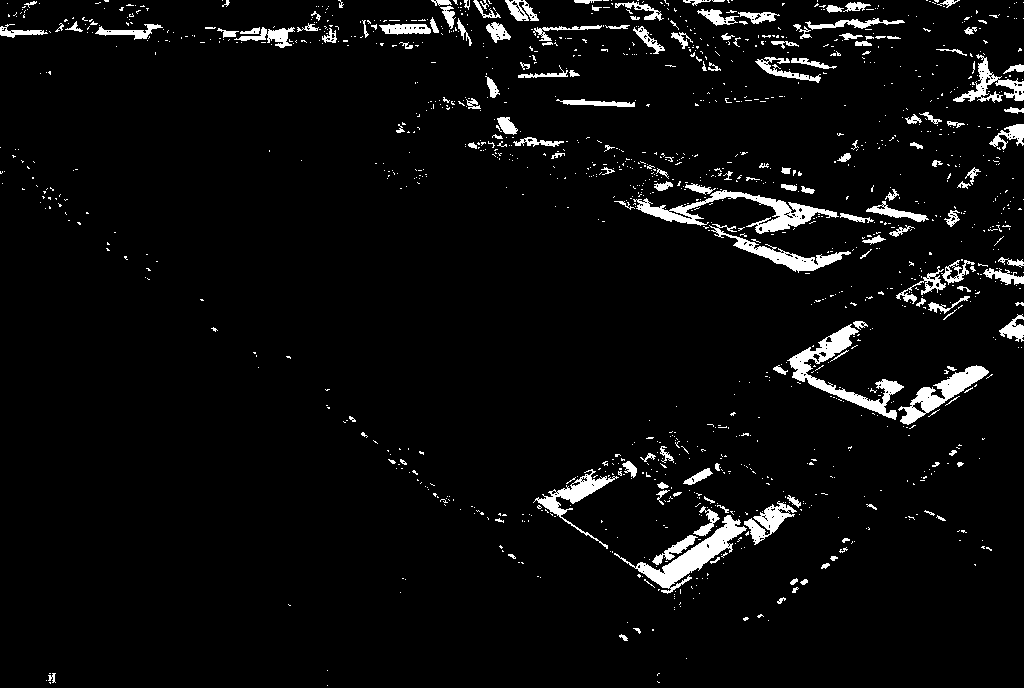}
        \caption{ }
        \label{subfig:vis4}
    \end{subfigure}
    \hfill
    \begin{subfigure}{.19\linewidth}
        \centering
        \includegraphics[width=\linewidth]{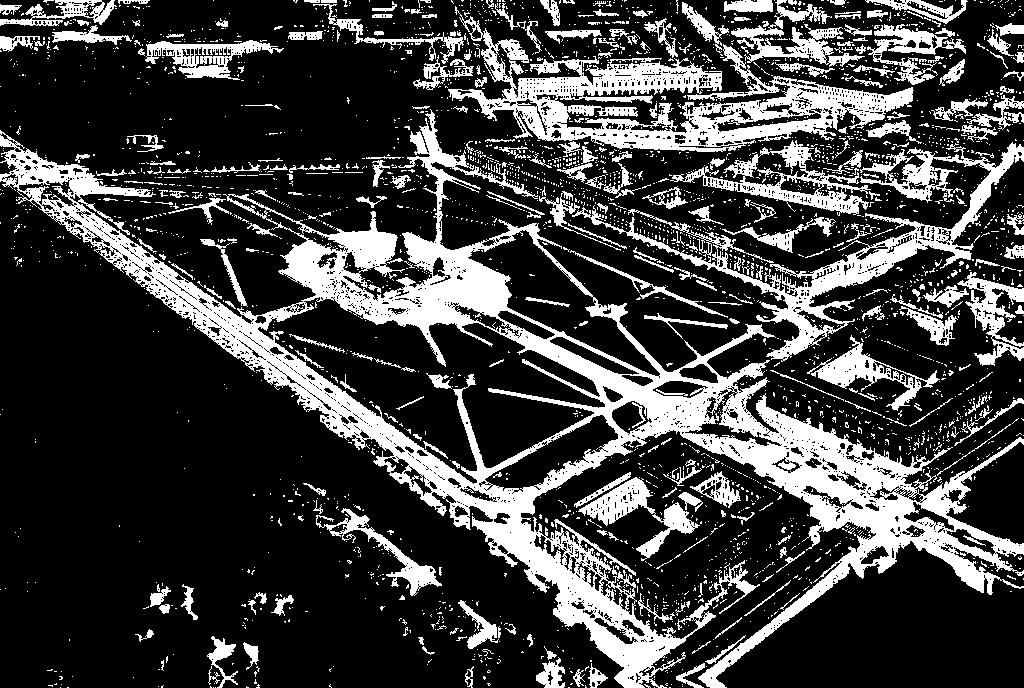}
        \caption{ }
        \label{subfig:vis5}
    \end{subfigure}
    \caption{Visualization of categorization results of adaptive category-based MSA. (a) is the input image. The white part of each binarized image from (b) - (e) represents a single attention category. }
    \label{fig:visualization of acmsa}
\end{figure*}

\subsection{Comparisons with State-of-the-Art Methods}
We choose the commonly used Set5~\cite{Bevilacqua2012set5}, Set14~\cite{Zeyde_2012_set14}, BSD100~\cite{Martin_2002_BSD100}, Urban100~\cite{Huang_2015_Urban100}, and Manga109~\cite{Matsui_2016_Manga109} as evaluation datasets and compare the proposed ATD model with current state-of-the-art SR methods.

We first compare our method with the state-of-the-art classical SR methods: EDSR~\cite{lim2017edsr}, RCAN~\cite{zhang2018rcan}, SAN~\cite{Dai_2020_san}, HAN~\cite{Niu_2020_han}, IPT~\cite{Chen_2020_ipt}, EDT~\cite{li2021efficient}, SwinIR~\cite{liang2021swinir}, CAT~\cite{chen2022cross}, ART~\cite{zhang2023accurate}, HAT~\cite{chen2023activating}. The results are presented in \cref{tab: results image sr1}
With comparable parameter size, the proposed ATD model significantly outperforms HAT~\cite{chen2023activating}.
Specifically, the ATD yields $0.20 - 0.25$ dB PSNR gains on the Urban100 dataset for different zooming factors.

For the lightweight SR task, we compare our method with CARN~\cite{Ahn_2018_carn}, IMDN~\cite{Hui_2019_imdn}, LAPAR~\cite{Li_2020_lapar}, LatticeNet~\cite{Luo_2020_latticenet}, SwinIR~\cite{liang2021swinir}, SwinIR-NG~\cite{Choi_2022_swinirng}, ELAN~\cite{zhang2022efficient}, and OmniSR~\cite{omni_sr}.
As can be found in \cref{tab: results image sr2} that under similar model size, the proposed ATD-light achieves better results with the recently proposed lightweight method OmniSR~\cite{omni_sr} on all benchmark datasets.
Our ATD-light outperforms OmniSR by a large margin (0.45dB) on the $\times4$ Manga109 benchmark.
Equipped with the token dictionary and category-based attention, our ATD-light model is able to make better use of external prior for recovering HR details under challenging conditions.

Extensive quantitative results have verified the efficacy of our ATD model.
To make qualitative comparisons, we provide some visual examples using different methods, as shown in \cref{fig:visualization 1}.
These images clearly demonstrate our advantage in recovering sharp edges and clean textures from severely degraded LR input.
For example, in img\_076 and MomoyamaHaikagura, most methods fail to reconstruct the correct shape, resulting in distorted outputs.
In contrast, our ATD can accurately recover clean edges with fewer artifacts, since it is capable of capturing similar textures from the entire image to supplement more global information.
More visual examples can be found in the supplementary material.

\subsection{Model Size and Computational Burden Analysis.}

\begin{table}
 \centering
 \caption{Model size and computational burden comparisons between ATD and recent state-of-the-art methods.}
 \label{tab:model size comparisons}
    \captionsetup{font={small}}
    \footnotesize
    \setlength{\tabcolsep}{4pt}

    \begin{tabular}{|l|cc|cc|cc|}
        \hline
        
        \multirow{2}{*}{\textbf{Model}} & \multirow{2}{*}{\textbf{Params}} & \multirow{2}{*}{\textbf{FLOPs}} & \multicolumn{2}{c|}{\textbf{Urban100}} & \multicolumn{2}{c|}{\textbf{Manga109}} \\
        & & & PSNR & SSIM & PSNR & SSIM \\
        \hline \hline
        CAT-A    & 16.6M & 360G & 27.89 & 0.8339 & 32.39 & 0.9285 \\
        HAT      & 20.8M & 412G & 27.97 & 0.9368 & 32.48 & 0.9292 \\
        ATD      & 20.3M & 417G & 28.17 & 0.8404 & 32.63 & 0.9306 \\
        \hline
    \end{tabular}
\end{table}

In this subsection, we analyze the model size of the proposed ATD model.
As shown in \cref{tab:model size comparisons}, we present the accuracy of image restoration (PSNR), model size (number of parameter) and computational burden (FLOPs) comparison between ATD and recent state-of-the-art models on image SR task.
Results in the table clearly demonstrate that the proposed ATD model helps the network achieve a better trade-off between restoration accuracy and model size.
Our ATD method achieves better SR results with comparable model size and complexity to HAT.
Furthermore, ATD outperforms CAT-A by up to 0.22 dB with only 10\% more parameters and FLOPs.

\subsection{Visualization Analysis}
We further visualize the categorization results in \cref{fig:visualization of acmsa} to verify the effectiveness of the category-based attention mechanism.
We use the binarized images to symbolize each attention category.
These illustrations clearly show that visually or semantically similar pixels are grouped together. 
Specifically, most of trees and shrubs are grouped in (b) and (c); the roof part is classified into (d), and (e) is dominated by the area of smooth texture in the image.
It indicates that the external prior knowledge of class information is incorporated into the token dictionary.
Therefore, AC-MSA can classify similar features into the same attention category, improving the accuracy of the attention map and performance.
This again confirms the rationality and effectiveness of category-based attention mechanism.

\section{Conclusion}
In this paper, we proposed a new Transformer-based super-resolution network.
Inspired by traditional dictionary learning methods, we proposed learning token dictionaries to provide external supplementary information to estimate the missing high-quality details.
We then proposed an adaptive dictionary refinement strategy which could utilize the similarity map of the preceding layer to refine the learned dictionary, allowing it to better fit the content of a specific input image.
Furthermore, with the external prior embedding in the token dictionary, we proposed to categorize input features and perform self-attention within each category.
This category-based attention transcends the limit of local window, establishing long-range connections between similar structures across the image.
We conducted ablation studies to demonstrate the effectiveness of the proposed token dictionary, adaptive refinement strategy, and adaptive category-based attention.
We have presented extensive experimental results on a variety of benchmark datasets, and our method has achieved state-of-the-art results on single image super-resolution. 

\vspace{7.5mm}

{
    \small
    \bibliographystyle{ieeenat_fullname}
    \bibliography{main}
}

\clearpage
\setcounter{section}{0}
\setcounter{table}{0}
\setcounter{figure}{0}
\maketitlesupplementary

\renewcommand\thesection{\Alph{section}}
\renewcommand\thetable{\Alph{section}.\arabic{table}}
\renewcommand\thefigure{\Alph{section}.\arabic{figure}}

In this supplementary material, we present more implementation details and additional visual results. We first provide training details of our ATD and ATD-light model in \cref{sec:training details}.
Then, we present more illustrations of AC-MSA and visual examples by different models in \cref{sec:more visual examples}.

\section{Training Details}
\label{sec:training details}
\paragraph{ATD.}
We follow previous works~\citep{liang2021swinir, chen2023activating} and choose DF2K (DIV2K~\citep{timofte2017div2k} + Flickr2K~\citep{lim2017edsr}) as the training dataset for ATD.
Then, we implement the training of ATD in two stages.
In the first stage, we randomly crop low-resolution (LR) patches of shape $64 \times 64$ and the corresponding high-resolution (HR) image patches for training.
The batch size is set to 32, while commonly used data augmentation tricks including random rotation and horizontal flipping are adopted in our training stage.
We adopt AdamW~\citep{loshchilov2018decoupled} optimizer with $\beta_1=0.9, \beta_2=0.9$ to minimize $L_1$ pixel loss between HR estimation and ground truth.
For the case of $\times 2$ zooming factor, we train the model from scratch for 300k iterations. The learning rate is initially set as $2\times 10^{-4}$ and halved at 250k iteration milestone.
In the second stage, we increase the patch size to $96\times 96$ for another 250k training iterations to better explore the potential of AC-MSA.
We initialize the learning rate as $2\times 10^{-4}$ and halve it at [150k, 200k, 225k, 240k] iteration milestones.
We omit the first stage for $\times 3$ and $\times 4$ models to save time, only adopting the second stage for finetuning these models based on the well-trained $\times 2$ model.
To ensure a smooth training process for the token dictionary, we set warm-up iterations at the beginning of each stage. During this period, the learning rate gradually increases from zero to the initial learning rate.

\paragraph{ATD-light.}
To make fair comparisons with previous SOTA methods, we only employ DIV2K as training dataset.
Same as ATD and previous methods, we train the $\times 2$ model from scratch and finetune the $\times 3$ and $\times 4$ models from the $\times 2$ one.
Specifically, we train the $\times 2$ ATD-light model for 500k iterations from scratch and finetune the $\times 3$, $\times 4$ model for 250k iterations based on the well-trained $\times 2$ model.
The larger patch size used for ATD is not applied to ATD-light.
The initial learning rate and iteration milestone for halving learning rate are set as $5\times 10^{-4}$, [250k, 400k, 450k, 475k, 490k] for $\times 2$ model and $2\times 10^{-4}$, [150k, 200k, 225k, 240k] for $\times3$, $\times 4$ models. The rest of the training settings are kept the same as ATD.

\section{More Visual Examples.}
\label{sec:more visual examples}

\begin{figure}
    \centering
    \captionsetup{font={small}}
    
    \includegraphics[width=0.9\linewidth]{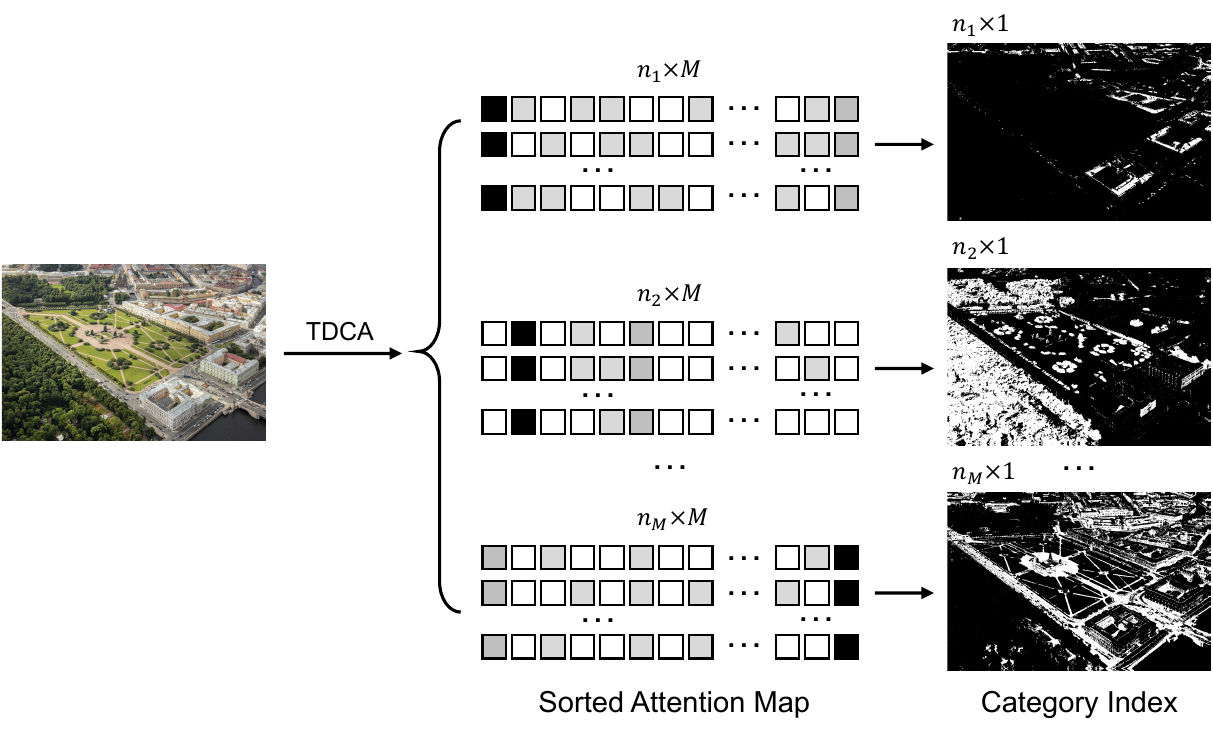}
    
    \vspace{-3mm}
    \caption{An illustration of the $\operatorname{Categorize}$ operation in \cref{eq:categorize}.
    With the attention map obtained by TDCA operation, we assign a category index to each pixel based on the highest similarity between the pixel and the token dictionary.
    }
    
    \label{fig:visualization of categorize}
    \vspace{-3mm}
\end{figure}

\subsection{More Visualization of AC-MSA.}
\begin{figure*}
    \centering
    \captionsetup{font={small}}
    
    \includegraphics[width=0.9\linewidth]{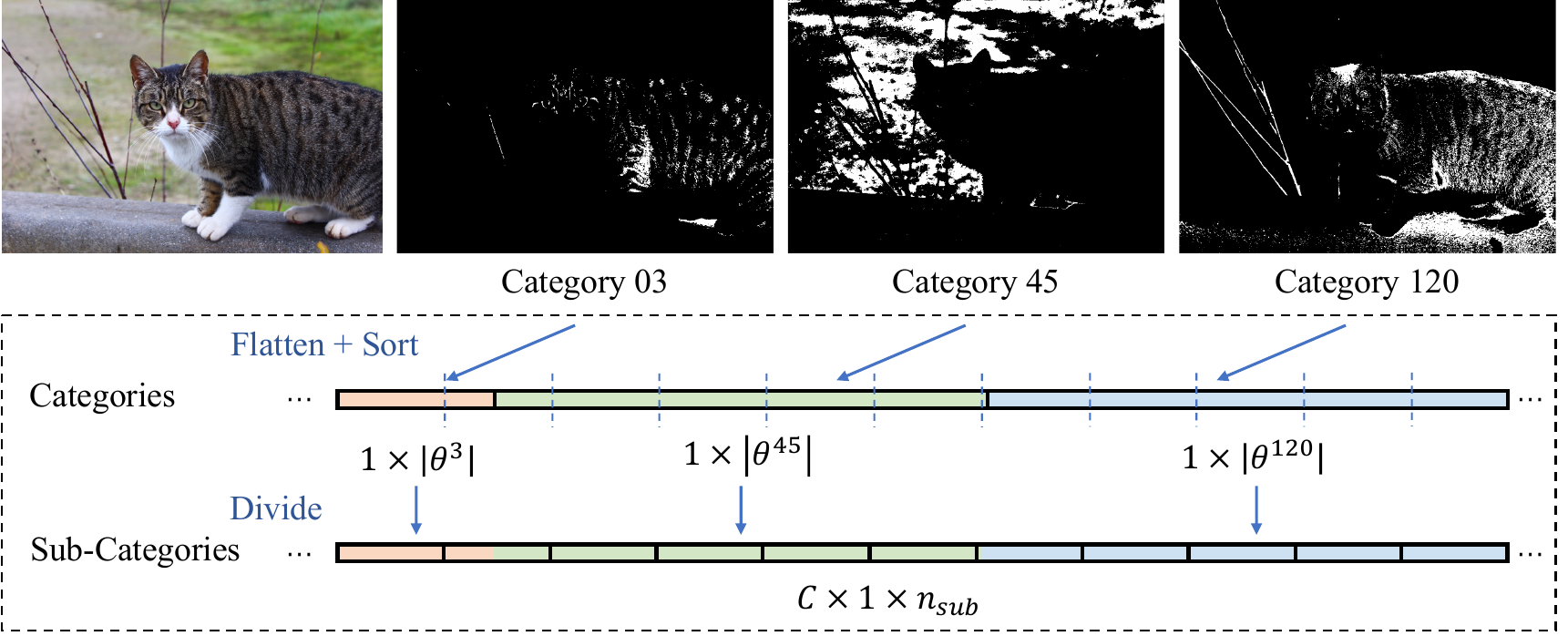}
    
    \vspace{2mm}
    
    \includegraphics[width=0.9\linewidth]{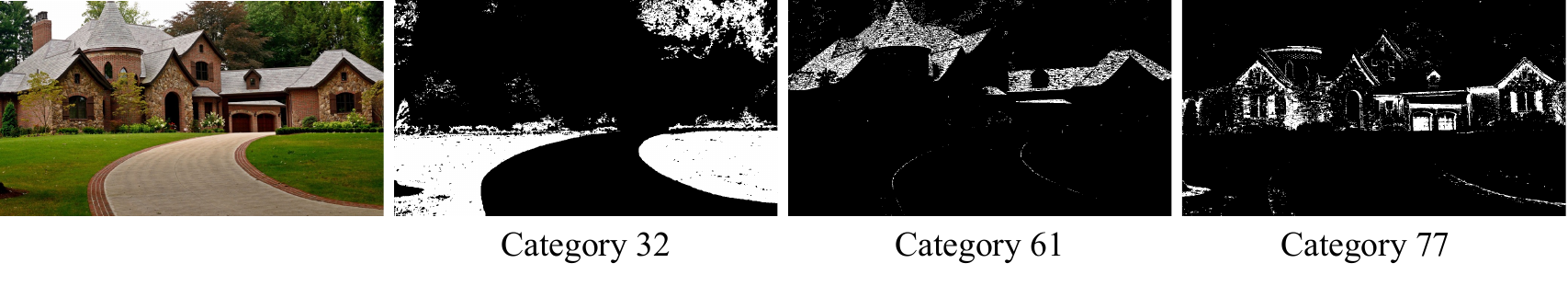}
    
    \includegraphics[width=0.9\linewidth]{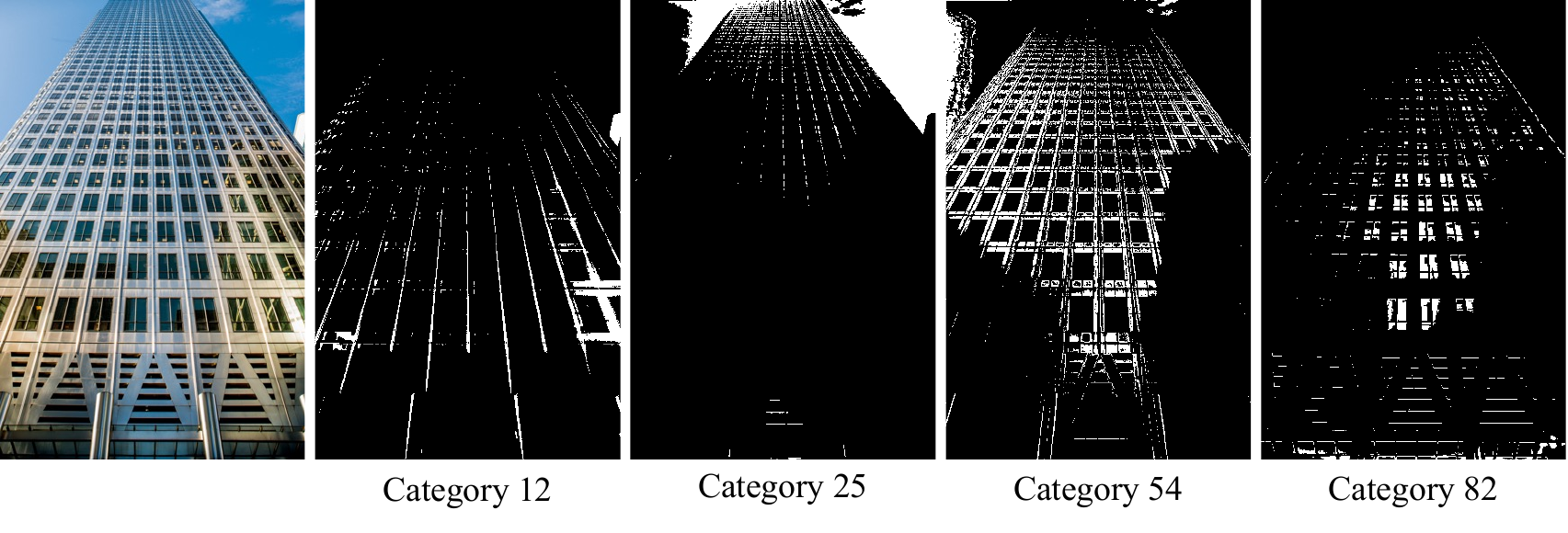}
    
    \includegraphics[width=0.9\linewidth]{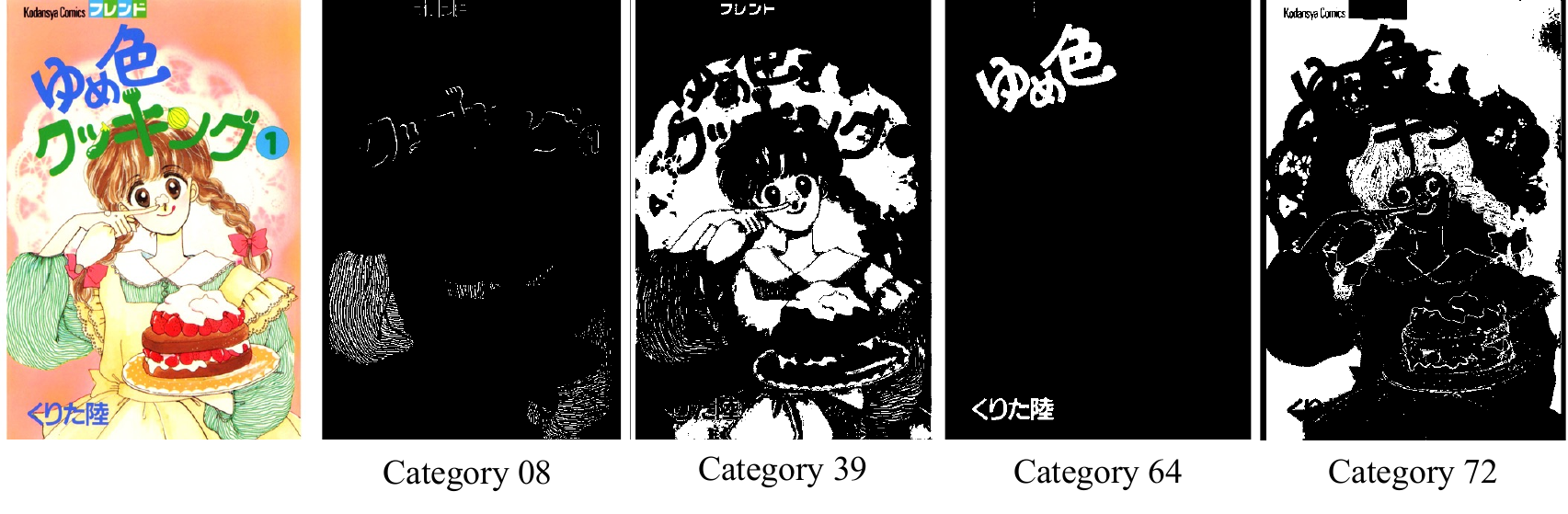}
    
    \vspace{-3mm}
    \caption{An illustration of the $\operatorname{Categorize}$ operation in \cref{eq:subcategorize}, along with several visual examples of the categorization results.
    The white area in each binarized image represents a single category. 
    Pixels in each category are flattened and then sorted for further dividing into sub-categories.
    These categorization results indicate that our AC-MSA is capable of dividing the image by the class of each pixel. Therefore, areas with similar texture (for example, sky, grass, roof) are grouped into the same category.
    }
    \label{fig:more visualization ac-msa}
    \vspace{-3mm}
\end{figure*}

In this subsection, we provide illustrations of the $\operatorname{Categorize}$ operation and more visual examples of categorization results in \cref{fig:visualization of categorize} and \cref{fig:more visualization ac-msa}.
We visualize only a few categories for each input image for simplicity.
In the $\operatorname{Categorize}$ operation, pixels are first sorted and classified into $\bm{\theta}^1, \bm{\theta}^2, \cdots, \bm{\theta}^M$ based on the value of attention map as in \cref{eq:categorize}. 
Then, each category is flattened and concatenated sequentially as in \cref{eq:subcategorize}.
Although certain pixels not belonging to the same category may be assigned to the same sub-category, it has almost no impact on performance.
This is because the number of misassignments will not exceed the dictionary size $M=128$, which is much less than the number of sub-categories $HW/n_s$.

The following visual examples further demonstrate that the categorize operation is capable of grouping similar textures together.
We can see that the categorize operation performs well on various types of image, including either natural or cartoon images.

\subsection{More Visual Comparisons.}
\begin{figure*}
    \centering
    
    \captionsetup{font={small}}
    
    \includegraphics[width=.85\linewidth]{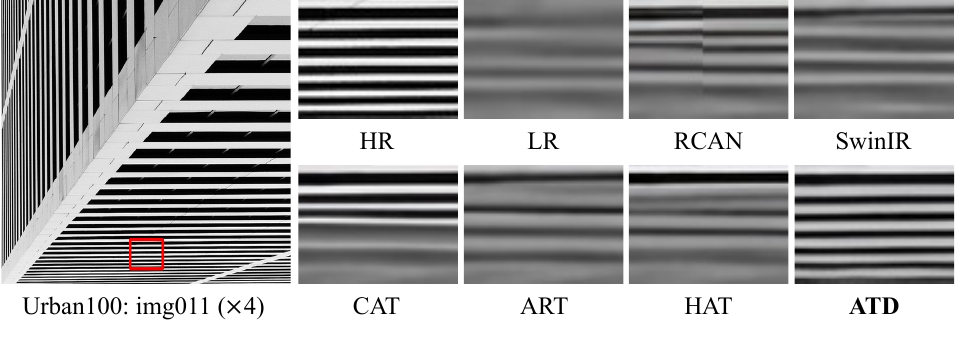}
    
    \includegraphics[width=.85\linewidth]{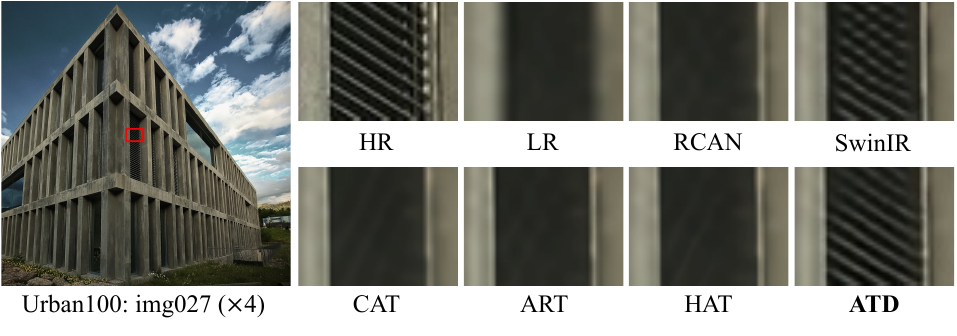}
    
    \includegraphics[width=.85\linewidth]{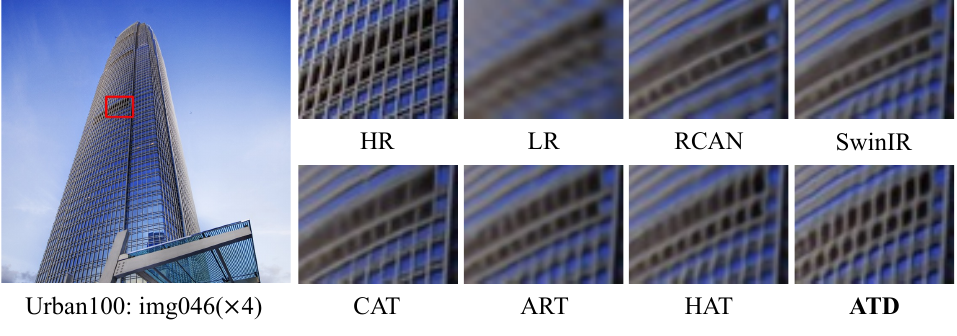}
    
    \includegraphics[width=.85\linewidth]{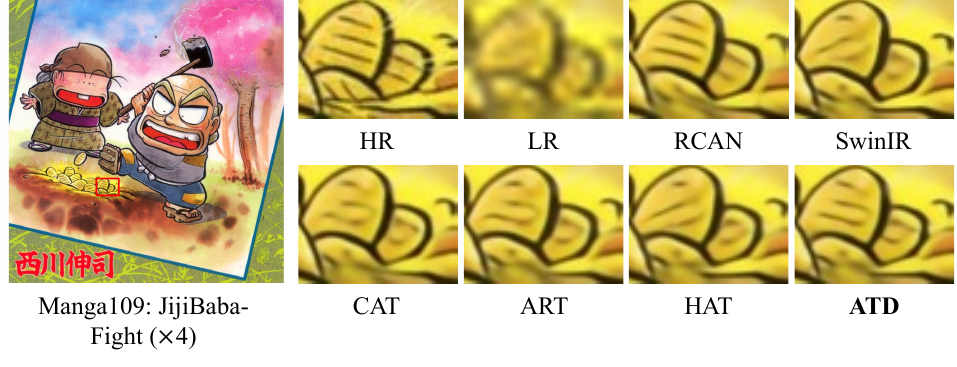}

    \vspace{-3mm}
    \caption{Visual comparisons between ATD and state-of-the-art classical SR methods.}
    \label{fig:more visual comp atd}
    \vspace{-3mm}
\end{figure*}

\begin{figure*}
    \centering
    
    \captionsetup{font={small}} 
    
    \includegraphics[width=.77\linewidth]{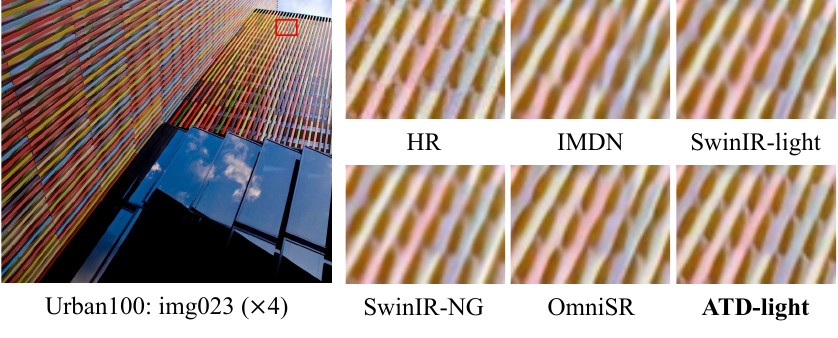}
    
    \includegraphics[width=.77\linewidth]{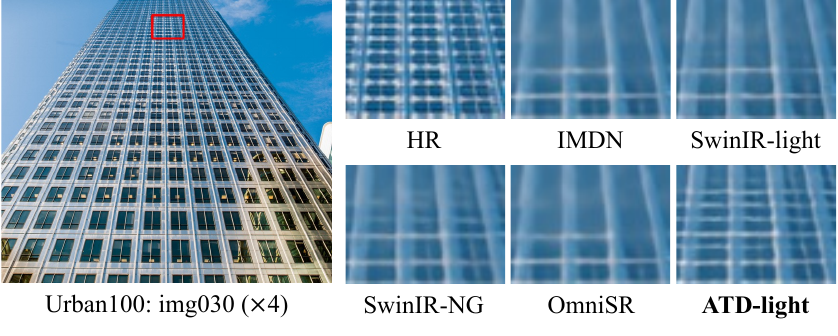}
    
    \includegraphics[width=.77\linewidth]{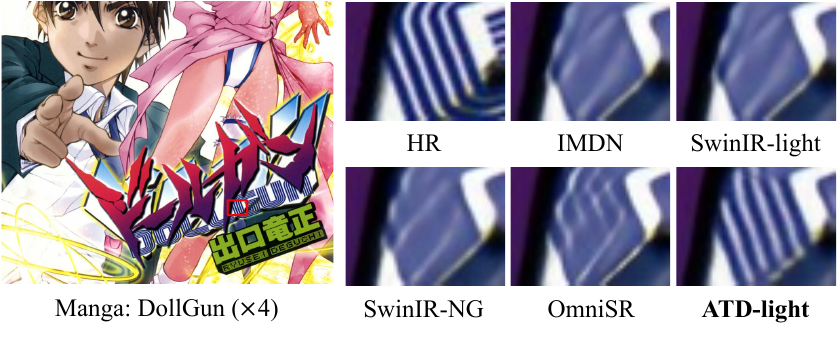}
    
    \includegraphics[width=.77\linewidth]{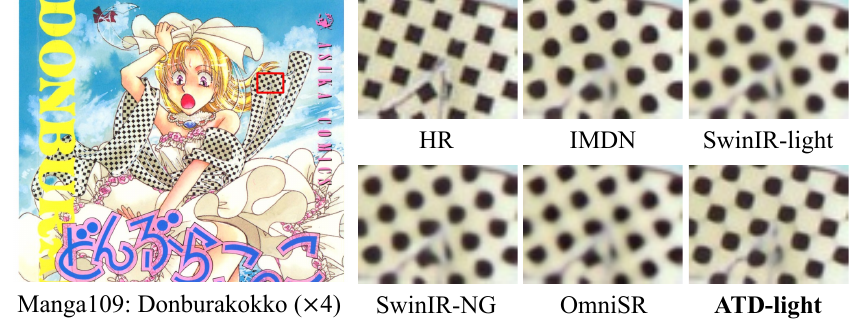}

    \vspace{-3mm}
    \caption{Visual comparisons between ATD-light and state-of-the-art lightweight SR methods.}
    \label{fig:more visual comp atdlight}
    \vspace{-3mm}
\end{figure*}

In this subsection, we provide more visual comparisons between our ATD models and state-of-the-art methods.
As shown in \cref{fig:more visual comp atd} and \cref{fig:more visual comp atdlight}, ATD and ATD-light both yield better visual results.
Specifically, ATD recovers sharper edges in img011 and img027, while the output of other methods remains blurry.
Moreover, most existing methods fail to reconstruct correct shape of the black blocks in Donburakokko.
In contrast, the output of ATD-light is more accurate to the rectangular shape in the ground truth.

\end{document}